\documentclass[letterpaper]{article} 
\usepackage[]{aaai25}  
\usepackage{times}  
\usepackage{helvet}  
\usepackage{courier}  
\usepackage{microtype}
\usepackage{pifont}
\usepackage{amsmath}
\usepackage{bbding}
\usepackage{multirow}

\usepackage{xcolor}

\usepackage[hyphens]{url}  
\usepackage{graphicx} 
\usepackage{natbib}  
\usepackage{caption} 
\usepackage{algorithm}
\usepackage{algorithmic}

\usepackage{newfloat}
\usepackage{listings}
\DeclareCaptionStyle{ruled}{labelfont=normalfont,labelsep=colon,strut=off} 
\floatstyle{ruled}
\newfloat{listing}{tb}{lst}{}
\floatname{listing}{Listing}
\title{Autoregressive + Chain of Thought $\simeq$ Recurrent:  \\ Recurrence's Role in Language Models' Computability and a Revisit of Recurrent Transformer}
\author{
  Xiang Zhang\textsuperscript{\rm 1},
  Muhammad Abdul-Mageed\textsuperscript{\rm 1,2},
  Laks V.S. Lakshmanan\textsuperscript{\rm 1}
}
\affiliations{
    \textsuperscript{\rm 1} University of British Columbia\\
    \textsuperscript{\rm 2} MBZUAI
}

\usepackage{bibentry}

\begin{document}

\maketitle

\begin{abstract}
The Transformer architecture excels in a variety of language modeling tasks, outperforming traditional neural architectures such as RNN and LSTM. This is partially due to its elimination of recurrent connections, which allows for parallel training and a smoother flow of gradients. However, this move away from recurrent structures places the Transformer model at the lower end of Chomsky's computational hierarchy, imposing limitations on its computational abilities. Consequently, even advanced Transformer-based models face considerable difficulties in tasks like counting, string reversal, and multiplication. These tasks, though seemingly elementary, require a level of computational complexity that exceeds the capabilities of the Transformer architecture.
Concurrently, the emergence of ``Chain of Thought" (CoT) prompting has enabled Transformer-based language models to tackle tasks that were previously impossible or poorly executed. Despite some previous research primarily interpreting CoT from a psychological perspective, a comprehensive understanding of \textit{why} CoT proves so effective in the reasoning process remains elusive. In this work, we thoroughly investigate the influence of recurrent structures in neural models on their reasoning abilities and computability, contrasting the role autoregression plays in the neural models' computational power. We then shed light on how the CoT approach can mimic recurrent computation and act as a bridge between autoregression and recurrence in the context of language models. It is this approximated recurrence that notably improves the model's performance and computational capacity. 
Moreover,  we revisit recent recurrent-based Transformer model designs, focusing on their computational abilities through our proposed concept of ``recurrence-completeness" and identify key theoretical limitations in models like Linear Transformer and RWKV.
Through this, we aim to provide insight into the neural model architectures and prompt better model design.
\end{abstract}

%

\section{Introduction}

\begin{figure}[htbp]
    \centering
    \includegraphics[width=0.8\columnwidth]{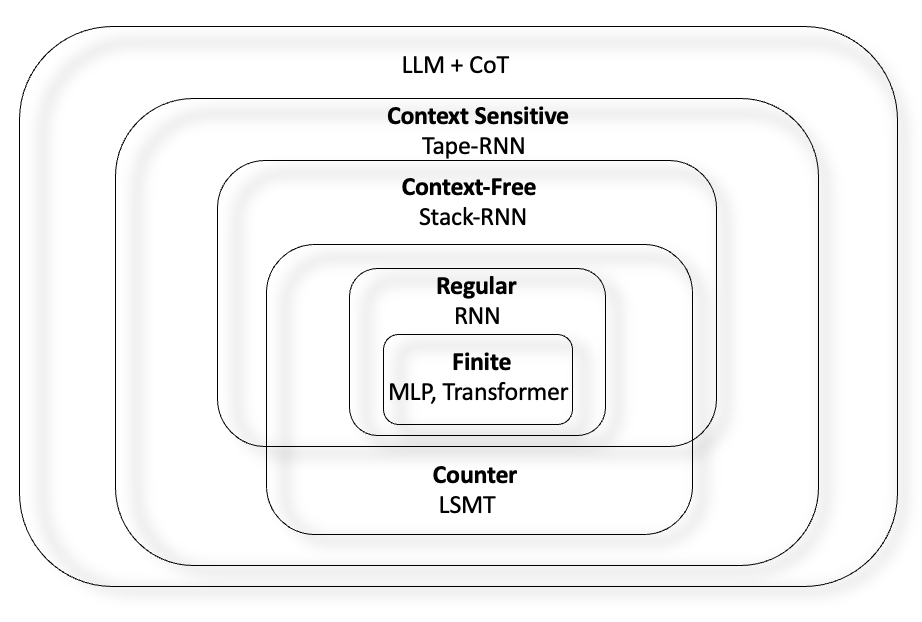}
    \caption{
    Computability hierarchy with each neural network architecture according to experimental results.
    }
    \label{fig:1}
\end{figure}
The emergence of large language models (LLMs) \cite{achiam2023gpt,touvron2023llama,jiang2023mistral}, featuring billions to trillions of parameters, marks significant progress in diverse language-related tasks \cite{chang2024survey,thirunavukarasu2023large,zhang2023don,wu2023bloomberggpt,beltagy2019scibert}. However, growing concerns have arisen over the limitations~\cite{dziri2024faith,valmeekam2022large,ullman2023large} of current LLMs, particularly their difficulties with basic tasks such as multiplication or counting. While much of the debate centers on training techniques and data choice~\cite{yu2024large}, it is crucial to also consider the theoretical limitations of the computational capabilities of these models, which fundamentally depend on their core architecture, Transformers~\cite{vaswani2017attention}.

In contrast to deterministic models like state machines or $K$ Nearest Neighbor classifiers, whose computational power is entirely reliant on their architectural (algorithm) design, the power of Neural Networks hinges upon a combination of both architecture~\cite{zhou2019analysis} and network optimization~\cite{deletang2022neural}. A Neural Network starting with random weights without any optimization, regardless of its architecture, has limited computational ability. Conversely, a network with a single linear layer without activation functions, even if perfectly optimized, is limited to capturing only basic linear relationships. 

Prior research~\cite{deletang2022neural,dziri2024faith,svete2024lower, chiang2301tighter} has demonstrated that recurrent neural networks~\cite{medsker2001recurrent} (RNNs and LSTMs) possess strong computational capabilities when optimally tuned, as supported by both empirical~\cite{deletang2022neural,dziri2024faith} and theoretical~\cite{svete2024lower, chiang2301tighter} studies. However, recurrent networks face significant optimization challenges~\cite{alqushaibi2020review}, such as the inability to parallelize during training and susceptibility to gradient vanishing~\cite{hochreiter1998vanishing} or exploding~\cite{kanai2017preventing} with longer sequences~\cite{ribeiro2020beyond}, which limits their scalability with large models and datasets. Conversely, the Transformer model replaces recurrence with an attention mechanism, enabling parallel training and mitigating the gradient vanishing issue through multiple gradient paths~\cite{abnar2020quantifying}. This innovation has made Transformers the leading choice for scalability~\cite{kaplan2020scaling} and optimization efficiency with large training data and model sizes. Nonetheless, the removal of recurrence imposes significant limitations on many reasoning tasks, as shown in multiple previous works~\cite{deletang2022neural,dziri2024faith}. Our work   further examines in depth the different roles of recurrence and autoregression in neural networks, revealing the necessity of recurrence for higher computational power.

\begin{figure}[htbp]
    \centering
    \includegraphics[width=1\columnwidth]{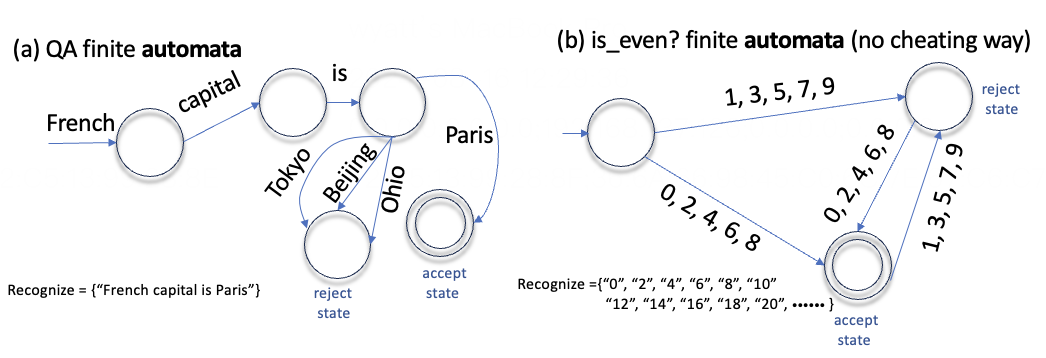}
    \caption{
    A comparison between using state machine to recognize the language of question answering and language of all the even numbers. Natural language tends to have wide branches but shallow depth whereas  logical reasoning can be very deep as input sequence can be arbitrarily long.
    }
    \label{fig:comp}
\end{figure}

The introduction of Chain of Thought (CoT) prompting~\cite{wei2022chain} represents a significant advancement for transformer-based language models, greatly enhancing performance across a range of tasks~\cite{chu2023survey}, including those beyond the computational capacity of the Transformer architecture. Despite substantial research analyzing the logic behind CoT, much of it interprets CoT from a psychological perspective~\cite{miao2024chain,li2024chain} as this way of reasoning is more human-like. Additionally, previous studies have examined CoT's role in knowledge extraction in LLMs~\cite{zhu2023physics}, which can differ from its role in \textit{reasoning} processes.
In this work, we elucidate CoT from a computability perspective, demonstrating that CoT approximates the omitted recurrence in the Transformer architecture. We show that CoT acts as a bridge between autoregression and recurrence, backing our claim with extensive experimental results and case studies involving tasks of varying computational levels.

Lastly, we revisit recent efforts to modify the Transformer architecture to be recurrent, including various architectural designs such as the universal Transformer~\cite{dehghani2018universal} and the linear Transformer~\cite{katharopoulos2020transformers}. However, we find that some of these so-called ``recurrent'' designs are primarily intended to reduce inference costs and do not enhance the model's computational power. We analyze the computational capabilities of each design and their ability to model recurrent functions through our proposed concept of \textit{Recurrence-Completeness}. Our analysis identifies key limitations of several recently proposed architectural modifications.

The contributions of this work are:
1) We define and contrast the concepts of recurrence and autoregression in neural networks, providing an in-depth analysis of their impact on a model's computational power.
2) We examine CoT from a computational standpoint, highlighting its role in bridging autoregression and recurrence in LLMs, supported by empirical evidence.
3) We systematically revisit and analyze recent recurrent modifications of Transformer architectures, uncovering the advantages and disadvantages of each design from a computational perspective.

\section{Definition and Concept}
Our study places significant emphasis on the concept of \textit{recurrence} within neural networks. However, the concepts of recurrence and autoregression have not been clearly defined and contrasted with in previous literature. In this section, we provide explicit definitions and distinctions between recurrence and autoregression in the context of neural networks to establish a foundation for our analysis.
\subsection{Recurrence and Autoregression}
A neural network can generate two types of outputs for a given input: (1) $\mathbf{h}$, representing the neural network's hidden state encoded as a vector, and (2) $\mathbf{o}$, the token (label), a single value derived from the hidden state vector $\mathbf{h}$. The use of these outputs shapes the network's modeling capabilities, resulting in either autoregressive or recurrent architectures.
 
 The notion of recurrence is derived from computation theory, where a model maps input data to corresponding output values, represented as \( f:\mathcal{X} \mapsto \mathcal{H} \). In line with computability theory conventions, we restrict $\mathcal{X}$ and $\mathcal{H}$ to countable sets, where all elements in $\mathcal{X}$ can be enumerated in a specific order as $(\mathbf{x}_1, \mathbf{x}_2, \mathbf{x}_3, \cdots)$. We denote the mapping of an element $\mathbf{x}_t$ at position $t$ as $\mathbf{h}_t = f(\mathbf{x}_t)$.  
 Function ${f}$ is said to be ($k$ terms) recurrent under function  \( {g}: \mathcal{H}^k \mapsto \mathcal{H} \),  if the output of $f$ on $x_t$ can be generated by applying $g$ to the $k$ preceding outputs of $f$  as follows:
\begin{equation}
    \mathbf{h}_t= f(\mathbf{x}_t) = g(\mathbf{h}_{t-1}, \mathbf{h}_{t-2}, \mathbf{h}_{t-3}, \cdots, \mathbf{h}_{t-k})  
    \label{eq:recur}
\end{equation}
Note that function $g$ is usually much simpler than $f$.

Take the Fibonacci sequence, \( {f}(n) = \frac{{\varphi^n - (1 - \varphi)^n}}{{\sqrt{5}}} \), as an example. It is (2 terms) recurrent under a simpler function \( g_{\text{add}} (\text{a}, \text{b}) =\text{a} + \text{b} \), where the $n$th result  can be derived from its two predecessors:
\begin{align}
\mathbf{h}_{n} &= {g}_{\text{add}} ( \mathbf{h}_{n-1}, \mathbf{h}_{n-2} ) = \mathbf{h}_{n-1} + \mathbf{h}_{n-2}
\end{align}
This captures the core principle of recurrent modeling: the current computational result, $\mathbf{h}_t$, can be derived solely using previous outcomes. This is possible because the preceding terms of $\mathbf{h}$ encapsulate \textbf{\textit{all}} essential computational information required for subsequent calculations and can thus be reused to resume the computation for obtaining the next $\mathbf{h}$. By employing the simple function $g$, the model utilizes the knowledge in the previous $\mathbf{h}$ terms and  avoids the need to directly apply the complex function ${f}$ to the entire input $\mathbf{x}_t$ and restart the entire computation anew.

In contrast,  autoregression~\cite{fuller1981properties},  established in statistical analysis, posits that the current \textit{observation} (through statistical sampling from underlying distribution $\mathbf{h}_t$) at time \( t \), denoted as \( \mathbf{o}_t \), can be inferred using previous observations:
\begin{equation}
    \mathbf{o}_t = {g}( \mathbf{o}_{t-1}, \mathbf{o}_{t-2}, \ldots, \mathbf{o}_{t-k})
\end{equation}
From a computational perspective, each result \( \mathbf{o}_t \) is fundamentally distinct from the computational state \( \mathbf{h}_t \). Termed as an `observation,' \( \mathbf{o}_t \) captures only a \textit{part} of the information in the complete computational state \(\mathbf{h}_t \). Consequently, \( \mathbf{o}_t \) \textit{may} lack essential information required for continuing the computation to generate the next output. For example, consider a function $f$ which determines if the $n$th Fibonacci number is greater than $1000$. Then the partial information $\mathbf{o}_n$ for the \( n \)th Fibonacci number, \( \mathbf{o}_n = \langle\texttt{whether } \mathbf{h}_n \texttt{ is greater than 1000}\rangle \), is insufficient to be used for computation of the \( (n+1) \)th result, and therefore might need to start the calculation anew from the beginning (i.e., from $\mathbf{x}_1$).

At this point, we can contextualize autoregression and recurrence within the framework of neural models. Recall that in these models, \( \mathbf{h}_t \) denotes the model's hidden state output  at time step \( t \), and \( \mathbf{o}_t \) is the word (or label) generated from \( \mathbf{h}_t \).  The vector $\mathbf{h}_t$ embodies the entirety of the computation of the model up to time \( t \), as it is where the neural model performs \textbf\textit{all} its reasoning and stores its intermediate information and memory. In contrast, the generated word \( \mathbf{o}_t \) is a discrete, \textbf{\textit{partial}} representation derived from \( \mathbf{h}_t \), capturing only a  part of the total computational information. Recurrent models utilize previous computational states to compute the current computational state \( \mathbf{h}_t \), as shown:
\begin{equation}
    \mathbf{h}_t = {g}_\theta(\mathbf{h}_{t-1:t-k})
\end{equation}
where ${g}_\theta$ is function represented by the neural model. By contrast, an autoregressive model  solely uses previously generated partial information (tokens) \( \mathbf{o}_{t-1:t-k} \) when calculating the current computational state. 
\begin{equation}
    \mathbf{h}_t = {g}_\theta(\mathbf{o}_{t-1:t-k})
\end{equation}
and further derives $\mathbf{o}_t$ from this $\mathbf{h}_t$ for future computation.

\subsection{Recurrence and Automata Theory }
Classic computational analysis is based on concecpt to Automata, the understanding of which is essential for applying computability theory to neural networks. A state machine, or Finite Automata (FA), is defined by a set of states and transitions governed by transition rules (Figure \ref{fig:comp}). Formally, FA is represented as a tuple \(\mathcal{M} = (\mathcal{Q}, \Sigma, \delta, q_0, \mathcal{F})\), where \(\mathcal{Q} = (q_1, q_2, \cdots, q_n)\) is a finite set of states, \(\Sigma\) is the input alphabet, \(\delta: \mathcal{Q} \times \Sigma \rightarrow \mathcal{Q}\) is the transition function, \(q_0\) is the initial state, and \(\mathcal{F}\) denotes the accepting states.

A FA generates two outputs upon processing an input $\mathbf{x}_t$: (1) $\mathbf{h}_t$, which is some resulting state $q_i$ from the set of all states 
$ \mathcal{Q}$; and  (2) $\mathbf{o}_t$,  a binary indicator,  $\texttt{True}$ or $\texttt{False}$, depending on whether the state $\mathbf{h}_t = q_i$ is an accepting or rejecting state. Similarly, state $q_i$ encapsulates all the information pertinent to the current state of the state machine, serving as the central hub for processing reasoning and cumulative memory. In contrast, $\mathbf{o}_t$ contains only partial binary information derived from $\mathbf{h}$ up to time $t$.

A state machine is inherently \textit{recurrent}, as the {transition function \( \delta \)} acts as the recurrent function \( g \) in  Equation \ref{eq:recur} on each output $\mathbf{h}$ of such machine. Specifically, \( \delta \) takes one previous computational state, i.e., \( \mathbf{h}_{t-1} = q_{j} \), and generates the current state, denoted as \( \mathbf{h}_t = q_i = \delta (\mathbf{h}_{t-1}, \sigma_t) \). Thus, for any input string \( \mathbf{x}_t  = \sigma_1 \cdots \sigma_t\), the final state \( \mathbf{h}_t = q_i \) is recurrently derived by applying \( \delta \) to the previous state \( \mathbf{h}_{t-i} = q_{j} \) for each input token, i.e., $\mathbf{h}_{t-i} = q_j = \delta(\mathbf{h}_{t-i-1}, \sigma_t)$.
For instance, when processing the string `1234' on the state machine depicted in Figure 2.b, the final state \( \mathbf{h}_{4} = q_{acc} \) is derived by applying the  transition function \( \delta \) recurrently to the computational result \( \mathbf{h}_{3} = q_{rej} \) of the preceding string `123'. And \( \mathbf{h}_{3} \) is obtained in the same way from its preceding string ``12", and so on. 
 Such recurrence is key to the power of a state machine, as computation can be continued for as many times as there are symbols in the input string $\mathbf{x}_t$, using the same function \( g(\cdot) = \delta(\cdot) \) and previous state $\mathbf{h}$. However, if we change the transitional function $\delta$ by having it take $\mathbf{o}_t \in $ $\{\texttt{True, False} \}$ as input, computation might not {progress correctly}   as $\texttt{True}$ or $\texttt{False}$ does not provide enough information for certain tasks to proceed to the next state when next token comes in.

\subsection{Time and Depth Complexity }
To illustrate the distinct roles of recurrence and autoregression within a given  neural model, we apply two complexity metrics in the reasoning process: time complexity and depth complexity. Time complexity measures the total computational operations executed to process an input of length $n$ utilizing the said model. In contrast, depth complexity measures the number of \textit{sequential} steps, after considering all parallel processing that a model performs, to process input $\mathbf{x}$. Depth complexity highlights the longest chain of dependent steps rather than the cumulative count of computational steps. Both complexities are quantified using  the  Big $O$ notation.

Different models exhibit varying complexities during the processing of inputs, based on their design. Nevertheless, each task has an inherent minimum complexity {(lower bound)} necessary for solving it. Models falling below this threshold are incapable of solving the task. For instance, multiplying two {$n$-bit} numbers requires a minimum of $\Omega(n\log n)$~\cite{afshani2019lower} time complexity, representing the total number of floating point operations needed  for an input of length $n$ and at least 
$O(\log n)$ depth complexity due to the possibility of parallelizing the multiplication of individual digits. The only sequential dependency arises in the subsequent addition of digits, which requires $\log n$ sequential steps if each pair of additions is performed simultaneously. Another example pertains to modeling a chess game with $n$ input moves, which requires  $O(n)$  depth complexity, as each board state calculation depends on both the current move and the previous state, and such dependency {does not admit} any parallelization. Models which exhibit lesser depth complexities for given input of length $n$, like Transformers, are thus ill-suited for, i.e., incapable of  tasks mentioned above, as we will show.

The computational complexity of a state machine is dictated by the number of times the transition function is invoked on the input.
As a state machine  is inherently recurrent, with each computation relying on sequential processing, contingent on prior states, its time and depth complexity are identical. In a deterministic finite state machine (DFA), both the depth and total computation precisely align with the length of the input string, resulting in a complexity of $O(n)$.

For a neural network, computation corresponds to each matrix multiplication $\mathbf{WX}$, with $\mathbf{W}$ being weight matrix and $\mathbf{X}$  the input. Even though each $\mathbf{WX}$ entails the summation of $n$ terms $\text{w}_1\text{x}_1 + \text{w}_2\text{x}_2 + \cdots + \text{w}_n\text{x}_n$, these summation operations are performed in  parallel, with no sequential dependencies. Consequently, the depth complexity  of each $\mathbf{WX}$ operation is $O(1)$.

\subsection{Memorization in Neural Network}

In practical applications, the effective depth \( c \) of matrix multiplication $\mathbf{WX}$ is not exactly $1$. This is due to large matrix multiplications combined with nonlinear functions, which can approximate complex functions and ``memorize" mapping results of computations requiring multiple sequential steps. For instance, results from multiplying large numbers can be memorized during training and retrieved via \( \mathbf{WX} \) in a single parallel computation, circumventing the typical dependencies. Hence, the constant \( c \) is proportional to the matrix size \( d \), denoted as:
\begin{equation}
    c \propto d = O(1)
\end{equation}

The size of matrix \( d \) in a neural network is influenced by the dimensionality of \( \mathbf{W} \) and the precision of its floating-point numbers. Increased dimensionality and precision allow for greater information storage. If precision were infinite, both the time and depth complexities of \( \mathbf{WX} \) could theoretically become infinite, transforming the matrix into a vast \textbf\textit{lookup table} through pure memorization.

However, with finite precision, merely storing the mapping results for specific tasks -- such as the outcomes of certain number multiplications -- does not truly ``solve" the task, as there exist larger input instances that exceed the matrix's memorization capacity. While memorization eliminates the necessity for recurrent computation and depth iteration for the memorized task instances, it often falls short in effectively solving tasks and demands exponentially more space.

\section{{CoT + Autoregressive $\simeq$ Recurrent}}

In this section, we show how recurrence enhances the computational capabilities of neural networks. 
A network with infinite precision ($d \rightarrow \infty$) could theoretically handle computations of infinite depth complexity by serving as a comprehensive lookup table for all feasible mappings. However, this ideal scenario is impractical. Motivated by this, our focus shifts to the practical setting of networks with finite precision. Furthermore, we explore the concept of chain-of-thought (CoT) \cite{wei2022chain} prompting as an approximation of recurrence within the domain of LLMs. Our analysis is situated within the realm of computability, delineating the upper bound of a model's computational capacity dictated by its architecture. We note that the impact of optimization techniques and training data on a model's computational capabilities falls beyond the scope of this work.

\subsection{Role of Recurrence in Computability}

To show the role of recurrence in neural models, we examine  the computational complexity exhibited by recurrent models (e.g., RNNs) as opposed to non-recurrent models (e.g., MLPs and  Transformers).

A Multi-Layer Perceptron (MLP)~\cite{popescu2009multilayer} is not recurrent, as it processes input of a fixed size and traverses through layers sequentially in a single iteration.  An MLP consists of $m$ layers, with each layer parameterized by  matrix $\mathbf{W}^{(i)}$, and performs matrix multiplication on the output from the previous layer $\mathbf{h}^{(i-1)}$:
\begin{equation}
    \mathbf{h}^{(i)} = \sigma( \mathbf{W}^{(i)}\mathbf{h}^{(i-1)} )
\end{equation}
where $\sigma$ denotes a non-linear function applied independently to each value of the resultant vector. This formulation diverges from the recursive definition in Equation \ref{eq:recur}, as each layer represents a distinct function  parameterized by different $W^{(i)}$ utilized only once in the forward pass rather than recurring upon itself.

As previously demonstrated, each \(\mathbf{WX}\) operation in an MLP entails a depth complexity of \(O(1)\), so an MLP with $m$ layers has a cumulative depth complexity of \(O(1) \times m = O(m)\) as layers are computed sequentially one after another. However, this complexity simplifies to \(O(1)\) because \(m\) remains constant for a \textit{given} MLP, irrespective of the input length $n$. This inherent limitation hinders MLPs from effectively addressing tasks like complex computations (e.g., multiplying large numbers) or string manipulations, requiring growing depth.

In contrast, an RNN~\cite{medsker2001recurrent} (with \(m\) layers) modifies an MLP  by integrating recurrent connections over the MLP itself. Specifically, the output from the last MLP layer, \(\mathbf{h}^{(m)}\), loops back as input for subsequent computations within the same RNN. Given an input sequence of \(n\) elements, denoted as \(\mathbf{x}_n = (\text{x}_1, \text{x}_2, \cdots, \text{x}_n)\), computations occur sequentially on each \(x_i\),  expressed at time $t$ as:
\begin{align}
    \mathbf{h}^{(m)}_t = \sigma(\mathbf{W}_{1}\mathbf{h}^{(m)}_{t-1} + \mathbf{W}_{2}\text{x}_{t})
\end{align}
This can be simplified to:
\begin{equation}
    \mathbf{h}^{(m)}_t = g_\theta(\mathbf{h}^{(m)}_{t-1}, \text{x}_t)
    \label{eq:rnn}
\end{equation}
Here, \(g_\theta\) signifies the function encapsulated by the RNN's MLP. This computation aligns with the definition of recurrence in Equation \ref{eq:recur} where the same model function $g_\theta$ iteratively operates upon itself. Given that each application of a given MLP represents a depth of \(O(m) = O(1)\), sequential application extends the depth complexity to \(O(n)\), with \(n\) indicating the input length.

\begin{table}[t]
\centering
\resizebox{0.8\linewidth}{!}{%
\begin{tabular}{c|c|c|c}

\hline
\multirow{2}{*}{\textbf{}} & \multirow{2}{*}{\textbf{Models}} & \textbf{Depth } & \textbf{Time}  \\

 &  & \textbf{Complexity} & \textbf{ Complexity}   \\

\hline\hline
\multirow{5}{*}{\shortstack{Models}} & DFA & O($n$) &O($n$) \\

 & MLP  & O($1$) & O($1$) \\

& RNN & O($n$) & O($n$) \\
& Transformers & O($1$) & O($n$)\\
& LLM + CoT & O($T(n)$) & O($n+T(n)$)\\
\hline



\hline

\end{tabular}}

\caption{Depth and time complexity of neural models with finite precision.  $T(n)$ denotes the Chain of Thought (CoT) steps for an input of length $n$.}
\label{tab:complexity}
\end{table}

The Transformer~\cite{vaswani2017attention}, despite its prowess in language modeling, does \textit{not} exhibit recurrent structure and has a limited depth. For an input sequence of length \( n \), the Transformer employs an attention mechanism which computes key (\( \mathbf{k} \)), query (\( \mathbf{q} \)), and value (\( \mathbf{v} \)) vectors for each input token \( x_i \) before they attend to each other for information retrieval. Specifically, at input step \( t \) and attention layer \( i \), the computations are as follows:
\begin{align}
    \mathbf{k}^{(i)}_t, \mathbf{q}^{(i)}_t, \mathbf{v}^{(i)}_t = \mathbf{W}_{k, q, v} \ \mathbf{h}_t^{(i-1)}\\
    \mathbf{h}_t^{(i)} =  \text{Attn} (\mathbf{k}^{(i)}_{1:t}, \mathbf{q}^{(i)}_t,  \mathbf{v}^{(i)}_{1:t})
     = \frac{\sum_{i=1}^{t} e^{\mathbf{q}_t^{(i)} \mathbf{k}_i^{(i)}} \mathbf{v}_i^{(i)}}{\sum_{i=1}^{t} e^{\mathbf{q}_t^{(i)} \mathbf{k}_i^{(i)}}}
     \label{eq:qkv}
\end{align} 
Since each \( \mathbf{k}, \mathbf{q} \), and \( \mathbf{v} \) is calculated from $\mathbf{h}^{(i-1)}$ at corresponding time $t$, we  can view the calculation of \( \mathbf{h}_t^{(i)} \) in Equation \ref{eq:qkv} as a function of all \( \mathbf{h}^{(i-1)} \) from step 1 to \( t \):
\begin{align}
    \mathbf{h}_t^{(i)} = g^{(i)}_\theta(\mathbf{h}^{(i-1)}_{1:t})
    \label{eq:transformer}
\end{align}
Here, \( g^{(i)}_\theta \) represents the function embodied by the $i$th attention layer. Unlike recurrent models, the output of each layer in the Transformer solely relies on the prior layer's output, devoid of self-referential loops. With a fixed number of layers \( m \), the computational steps remain limited to \( m \) sequentially executed stages, \textit{unaffected by input length} expansion. The final layer output $\mathbf{h}^{(m)}_t$ is only a function of the \textit{input} instead of the \textit{previous hidden state} (Figure \ref{fig:recurTransformer}a):
\begin{equation}
    \mathbf{h}^{(m)}_{t} = g_\theta(\text{x}_{1:t})
\end{equation}
Hence, the depth complexity is constrained to be \( O(1) \) by the fixed layer count.

A comparison between recurrent and non-recurrent models in Table~\ref{tab:complexity} underscores the pivotal role of recurrence in enhancing the depth of reasoning. This amplification is crucial for tackling tasks that demand growing depth during reasoning.

\subsection{Role of Autoregressive in Computability}
As demonstrated previously, autoregression is not a  substitute for recurrence in the computational process. Unlike a recurrent process, where the computed state \(\mathbf{h}\) is re-entered into the model as input, an autoregressive model condenses the entire computational state \(\mathbf{h}\) into a single token \(\mathbf{o}\) and  augments the input with \(\mathbf{o}\). For example, 
consider simulating a chess game with a sequence of \(n\) actions \(\mathbf{x}_n = (\text{x}_1, \text{x}_2, \cdots, \text{x}_n)\). The computational state \(\mathbf{h}\) must encode the board information at each step to avoid having to resort to memorization. An autoregressive model does not pass this calculated hidden state \(\mathbf{h}\) into the next calculation. Instead, the next chess move \(\mathbf{o}\) is derived from \(\mathbf{h}_t\), and this token \(\mathbf{o}\) is reintroduced into the model, resulting in a new input augmented sequence \(\mathbf{x}_{n+1} = (\text{x}_1, \text{x}_2, \cdots, \text{x}_n, \mathbf{o}_1)\). That is, the  autoregressive process extends the input sequence by appending the newly derived token to the original input. However, this does not enhance depth complexity since it does not alter the model structure but only the input. Because the tokens \(\mathbf{o}\) generally do not encode enough computational information, reasoning for the next move \(\mathbf{o}_{2}\) must start from scratch, unlike leveraging \(\mathbf{h}_t\) from the previous step in a recurrent process.

Therefore, autoregression preserves the original model’s depth complexity while increasing the time complexity as more computations are performed on the extended input.

\begin{figure*}[t!]
    \centering
    \includegraphics[width=0.7\linewidth]{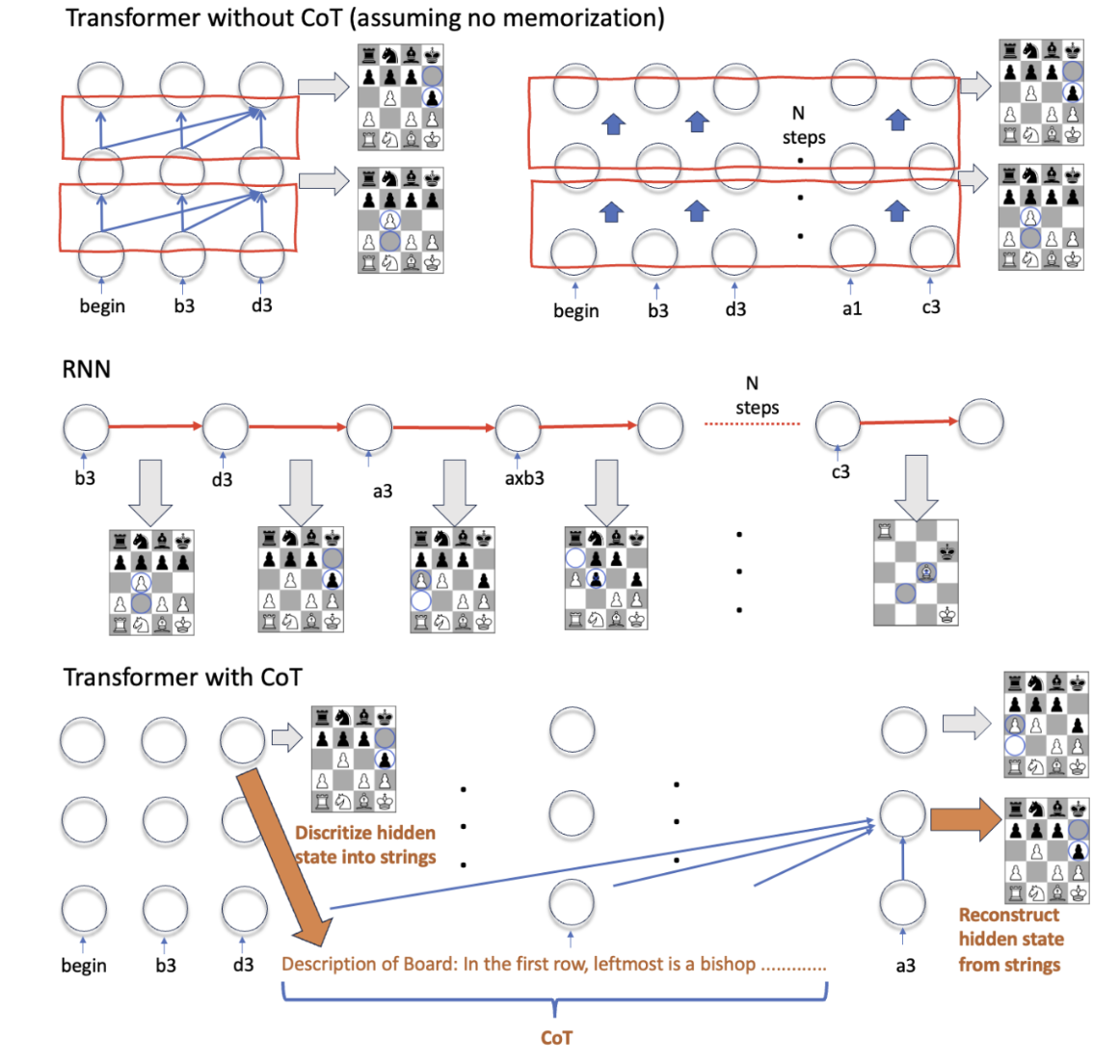} 
    \caption{Visualization of how computational information is passed along sequentially. Information between red colors is sequential (Between layers for transformers and across steps for RNN). Transformer without CoT can only pass the information through layers sequentially and therefore its depth is limited to layer numbers. RNN is recurrent over time therefore can pass the hidden information as many times as input length. CoT converts the hidden information from vectors into strings and then converts it back to vectors, therefore achieving approximate recurrence.}
    
    \label{fig:recurrence}
\end{figure*}

\subsection{Role of CoT in Computability}

Large language models (LLMs) are autoregressive models that utilize condensed outputs, \( \mathbf{o}_t \), derived from hidden states \( \mathbf{h}_t \) for subsequent computations. Natural language is a powerful medium for encoding various kinds of information. Specifically, the Chain of Thought (CoT) approach employs a sequence of natural language tokens, (\( \mathbf{o}_{1}, \mathbf{o}_{2}, \mathbf{o}_{3}, \ldots, \mathbf{o}_{k} \)), to form sentences that encode intermediate computational information from the hidden state \( \mathbf{h} \). This behavior is represented as \( \mathbf{h}^{(m)}_t \rightarrow \mathbf{o}_{1:k} \), where \(``\rightarrow"\) denotes discretizing and encoding the computation state information into string format.

In subsequent computations, instead of  solely using the task-related input \( \mathbf{x_n} = (\text{x}_1, \cdots, \text{x}_n) \), the encoded CoT strings are appended to form a new input \( \mathbf{x_{n+k}} = (\text{x}_1, \cdots, \text{x}_n, \mathbf{o}_1, \mathbf{o}_2, \cdots, \mathbf{o}_k) \). When this input is fed to the model, \( \mathbf{o}_{1:k} \), which encodes \( \mathbf{h}^{(m)}_t \), is converted back to the hidden vector, denoted as \( \mathbf{o}_{1:k} \rightarrow \mathbf{h}^{(1)}_{n+k} \). Since this string encodes the computational state represented by \( \mathbf{h}^{(m)}_t \), converting it back to the hidden state allows the model to directly utilize it to continue the computation from the recovered \( \mathbf{h}^{(m)}_t \), rather than reasoning from the beginning. The entire CoT process can be  represented as:
\begin{equation}
   \mathbf{h}^{(m)}_{n} \rightarrow \mathbf{o}_{1:k} \rightarrow \mathbf{h}^{(1)}_{n+k}
\end{equation}

Thus, the autoregressive process in CoT simulates the missing recurrent connection by iteratively encoding the computation state into strings and decoding the strings back to the computation state. Assuming CoT performs \( T(n) \) steps of the above conversion for a given task instance \( x_n \), the time complexity is enhanced to \( O(n + T(n)) \) through the autoregressive process during CoT, with a depth complexity of \( O(T(n)) \) attained through simulated recurrence from such string-vector conversions.

For example, in a chess-playing scenario, the computational state \( \mathbf{h} \), which encodes the board information, is converted to descriptive strings detailing the current chessboard configuration in the CoT process. Contrast that with directly outputting the next action as done by a non-CoT-based inference. The board description $\mathbf{o}_{1:k}$ can then be used by the CoT-based model to resume the computation, bypassing the need to compute from scratch by using only previous actions as input. An illustration of this process and how CoT approximates recurrent connections like RNN is shown in Figure \ref{fig:recurrence}.

\subsection{Role of CoT Variants in Computability}

As the naive CoT simply uses the prompt "Think Step by Step" and guides the model to output the reasoning state \(\mathbf{h}_t\) into a sequence of natural language tokens (\(\mathbf{o}_1, \mathbf{o}_2, \cdots, \mathbf{o}_k\)), it might not effectively convert all useful computational information from \(\mathbf{h}\) to \(\mathbf{o}_{1:k}\). Therefore, different CoT variants have been proposed, and we discuss how these different CoT methods affect the reasoning process and the model's computability.

\textbf{Tree of Thought (ToT).} Instead of outputting a single reasoning sequence \(\mathbf{o}_{1:k}\), ToT encourages the model to output multiple possible reasoning paths simultaneously. We denote the \(i\)-th reasoning path as \(\mathbf{o}_{1:k_i}^{(i)}\). Each path describes a different possible CoT reasoning logic to solve the problem. Then, we evaluate each one of them before expanding the reasoning on the most promising top \(L\) paths independently. Similarly, all reasoning paths are obtained through \(\mathbf{h}^{(m)}_t\) and this can be represented as:
\begin{equation}
\mathbf{h}^{(m)}_{n} \rightarrow (\mathbf{o}^{(1)}_{1:k_1}, \mathbf{o}^{(2)}_{1:k_2}, \cdots, \mathbf{o}^{(L)}_{1:k_L}) \rightarrow (\mathbf{h}^{(1)}_{n+k_1}, \cdots, \mathbf{h}^{(L)}_{n+k_L})
\end{equation}
As we can see, even though there might be \(L\) different reasoning paths, each path performs reasoning steps independently and conforms to our previous analysis of CoT. Each path extracts different reasoning (computational) information from \(\mathbf{h}\) and then discretizes the hidden computation into strings before converting these strings back to \(\mathbf{h}\). During this process, each path approximates recurrence on its own. Assuming the longest reasoning path in ToT performs \(T(n)\) steps of CoT, the depth complexity of ToT will be \(n + T(n)\), the same as CoT. Therefore, ToT does not increase the depth complexity beyond that of CoT but improves the conversion of \(\mathbf{h} \rightarrow \mathbf{o}\) by encoding multiple possible reasoning solutions.

Since \(\mathbf{h}\) cannot be directly passed to the next step as in a recurrent model, ToT explicitly extracts all possible solutions encoded in \(\mathbf{h}\) and further expands on them. For complex tasks, this can be helpful as some require searching rather than simple one-directional reasoning, and a single reasoning path \(\mathbf{o}_{1:k}\) might not encode all necessary information from $\mathbf{h}$ for continued computation. In such cases, naive CoT does not extract all necessary information from \(\mathbf{h}\) and therefore does not approximate the desired recurrence.

\textbf{Graph of Thought (GoT):} GoT enhances the Tree of Thought by introducing an iterative self-refinement and aggregation process. In ToT, each thought in the tree independently performs reasoning. In contrast, GoT merges the reasoning paths of these thoughts into a single unified path, allowing them to share reasoning information across paths rather than relying solely on their own. Additionally, GoT incorporates a self-refinement mechanism that evaluates its reasoning and makes corrections. These enhancements enable GoT to better extract correct and useful information from the underlying reasoning state, $\mathbf{h}$.

In summary, all variants of Chain of Thought (CoT) improve the process of transitioning from $\mathbf{h} \rightarrow \mathbf{o}_{1:k}$ for approximated recurrence. Since $\mathbf{h}$ contains a vast amount of information and computational intermediates, a simple CoT might struggle to extract the most useful elements (e.g., multiple solutions embedded in $\mathbf{h}$). Different CoT variants provide more effective ways to convert the hidden state into informative outputs. However, these variants do not enhance the process of $\mathbf{o} \rightarrow \mathbf{h}$, as encoding text into a hidden state is optimized during training. Additionally, variants of CoT do not further increase the depth complexity beyond what CoT already achieves as the total depth is decided by vector-string conversion steps $T(n)$.

\subsection{\texttt{Autoregressive + CoT $\simeq$ Recurrent} Holds Only in Language Models}

An implicit prerequisite for mimicking recurrence using Chain of Thought is that the tokens $\mathbf{o}_{1:k}$ must be expressive and universal enough to encode all types of information, including reasoning states, state memories, and intermediate computational results. Natural language is posited to be powerful enough to encode all sorts of information using natural language tokens. From chess boards and programs to data structures and computational graphs, strings can effectively encode them all in meaningful values. 

However, this does not hold true for certain non-natural language-based large models. For instance, protein language models that use 20 amino acids as tokens~\cite{lv2024prollama} cannot effectively convert hidden representations $\mathbf{h}$ into meaningful representations with amino acid tokens, as these tokens can only encode limited, rather than universal, information. Similarly, a pretrained chess model cannot perform autoregressive-based recurrent reasoning because it only has tokens representing chess moves, lacking the ability to convert $\mathbf{h}$ into descriptions of the chessboard.

An illustrative demonstration of how CoT achieves RNN-like recurrence is shown in Figure~\ref{fig:recurrence}.

\section{Experiments}
While we have highlighted the critical role of recurrence and the mechanisms of Chain of Thought (CoT), quantifying the Chomsky hierarchy-aligned computability of CoT-enhanced LLMs remains challenging. This involves analyzing memory structures beyond the depth complexity previously discussed. To empirically demonstrate the computational power of recurrence (both True and Approximate), we conduct experiments following  previous work~\cite{deletang2022neural} on examining different models' capability to solve tasks at each level. Table \ref{tab:performance} shows the results of our experiments, described next.

\subsection{Experiment Design}
\textbf{Model Choice.} \textit{The goal of our work and experiments is not to evaluate and compare the performance of different LLMs.} Instead, our aim is to demonstrate the role of CoT in approximating recurrence and show the improved computational power of incorporating recurrence. Specifically, our work focuses on the upper limit of the model's computational power based on architectural designs. Other factors such as optimization, training efficacy, and tokenizer choices are beyond the scope of this investigation. Therefore, we choose the best-performing model available to us, GPT-4~\cite{achiam2023gpt}, to cater to this purpose. Detailed model usage and prompt examples are shown in the Appendix.

\noindent\textbf{Tasks Choice.}  We follow previous work on the empirical analysis of the expressiveness of neural expert models~\cite{deletang2022neural} and adopt their task settings. Tasks are divided into three computational levels: Regular (R), which requires machines equivalent to or more powerful than a DFA; Context-Free (CF), solvable by Pushdown Automaton (PDA); and Context-Sensitive (CS), requiring linear-bounded Automaton (LBA).

Task input format can significantly influence LLM performance. For example, LLMs often mistakenly infer "9.9 $<$ 9.11" or count characters incorrectly due to suboptimal splitting during text tokenizing. To minimize these effects, we redesigned the tasks. Task instances like "aababababa" for string reversing are replaced with list reversing, e.g.,  of ["apple", "monkey", "apple", $\cdots$], as word like "apple" remains a single token in modern tokenizers. We also limit task length to avoid issues with long context access and cross-session problems in prompting. Detailed task designs, length sampling, and example inputs/outputs are given in the Appendix.

\begin{table*}[ht]
\centering
\resizebox{0.8\textwidth}{!}{
\begin{tabular}{lllllll|ll}
\hline
\textbf{Level} & \textbf{Task}                  & \textbf{RNN}   & \textbf{Stack-RNN} & \textbf{Tape-RNN} & \textbf{Transformer} & \textbf{LSTM}  & \textbf{LLM}  & \textbf{CoT}   \\ \hline\hline
R     
      & Modular Arithmetic    & \textbf{100.0} & \textbf{100.0}     & \textbf{100.0}    & 24.2        & \textbf{100.0} &    18.0   &   \textbf{100.0}     \\
      & Parity Check        & \textbf{100.0} & \textbf{100.0}     & \textbf{100.0}    & 52.0        & \textbf{100.0} &   48.0   &    \textbf{92.0}   \\
      & Cycle Navigation     & \textbf{100.0} & \textbf{100.0}     & \textbf{100.0}    & 61.9        & \textbf{100.0} &   24.0    &      \textbf{100.0}  \\ \hline
CF   & Stack Manipulation    & 56.0  & \textbf{100.0}     & \textbf{100.0}    & 57.5        & 59.1  &   0.0    &      \textbf{100.0}  \\
      & Reverse List       & 62.0  & \textbf{100.0}     & \textbf{100.0}    & 62.3        & 60.9  &    0.0   &    88.0    \\
      & Modular Arithmetic    & 41.3  & \textbf{96.1}      & \textbf{95.4}     & 32.5        & 59.2  &    0.0   &     \textbf{94.0}   \\\hline
CS 
      & Odds First            & 51.0  & 51.9      & \textbf{100.0}    & 52.8        & 55.6  &    0.0   &   \textbf{100.0}   \\
      & Addition       & 50.3  & 52.7      & \textbf{100.0}    & 54.3        & 55.5  &   0.0    &    \textbf{100.0}    \\
      & Multiplication & 50.0  & 52.7      & 58.5     & 52.2        & 53.1  &     0.0  &   56.0       \\
      & Sorting          & 27.9  & 78.1      & 70.7     & \textbf{91.9}        & \textbf{99.3}  &    0.0   &    \textbf{100.0}    \\ \hline
\end{tabular}}
\caption{Empirical results of each architecture's performance for different levels of tasks.}
\label{tab:performance}
\end{table*}

\subsection{Results}
The experiment results for LLM without  and with CoT with are appended to the expert model's performance from previous work~\cite{deletang2022neural} in Table~\ref{tab:performance}. 
As we can see, all recurrence-augmented models can solve tasks in the regular (R) category. This includes true recurrence models such as RNN, Stack-RNN~\cite{joulin2015inferring}, Tape-RNN~\cite{deletang2022neural}, and LSTM, as well as approximated-recurrence using CoT-based LLMs. Specifically, the accuracy for R tasks is nearly 100\% for all such models. In comparison, non-recurrent models, whether expert (trained for a specific task) or general-purpose LLM, struggle with R tasks. The accuracies on R tasks for Transformer expert models are far from ideal (20-60\% accuracy) compared to RNN (100\%), with Transformer-based LLMs (without CoT) performing even worse.

This further solidifies the complexity analysis shown in Table \ref{tab:complexity}. Specifically, all recurrent-based models, including CoT, possess a depth complexity greater than DFA, which is the minimum capability required for solving R tasks. However, since Transformer's depth complexity is constrained to be \(O(1)\), solving these tasks is infeasible.

Furthermore, CF and CS tasks require memory structures corresponding to a stack in PDA and a linear tape in a LBA, respectively. Not surprisingly, augmenting RNNs with the corresponding memory achieves high accuracy in each task level. However, since Transformer-based models have a depth complexity of \(O(1)\), even though their attention module allows for complex memory access, their limited reasoning depth prevents them from successfully solving tasks at each level. Specifically, we see Transformer expert models achieve low accuracy (30\%-60\%) in both CF and CS tasks. For non-expert LLMs without CoT, large failures are witnessed in solving any of these tasks, with an accuracy of 0\% for every single task in those categories.

However, this inability to model higher-level complexity is mitigated when CoT is introduced. As seen in Table \ref{tab:performance}, augmenting LLMs with CoT significantly improves testing accuracy on CF and CS tasks. Except for list reversing and multiplication, where accuracy falls below 90\%, performance on other tasks is close to 100\%. Even though LLMs are not augmented with specific memory structures, the CoT process can intuitively act as a storage medium using the output text. Transformer-based LLMs can achieve tape-like memory random access through their attention mechanism on the CoT-generated text. In summary, CoT augments LLMs with the depth complexity required for solving all levels of tasks. In the Appendix, we further illustrate this recurrence approximation with extensive case studies on the output from LLMs.

\begin{table}[t]
\centering
\resizebox{0.7 \linewidth}{!}{%
\begin{tabular}{c|c|c|c}
\hline

\textbf{Transformer} & \multirow{2}{*}{\textbf{Depth}} &  \multirow{2}{*}{\textbf{PT}}  & \textbf{Recurrence}   \\ 

\textbf{Type} &   & &\textbf{Completeness} \\\hline
\hline
Standard Recurrent & \(O(\text{n})\) & \XSolidBrush &Complete \\\hline
FeedBack Recurrent & \(O(\text{n})\) & \XSolidBrush & Complete \\\hline
Block Recurrent  & \multirow{2}{*}{\(O(\text{n} / k)\)}  & \multirow{2}{*}{\XSolidBrush} & \multirow{2}{*}{Complete}  \\
(block size = k) & & &  \\ \hline
RWKV & \(O (1)\)  & \Checkmark & Incomplete \\ \hline
Linear Transformer & \(O (1)\)  & \Checkmark & Incomplete \\ \hline

 \hline
\end{tabular}}
\caption{All types of Recurrent Transformers and their properities. PT stands for parallel training.}\label{tab:recTsf}
\end{table}

\begin{figure*}[htbp]
    \centering
    \includegraphics[width=1\linewidth]{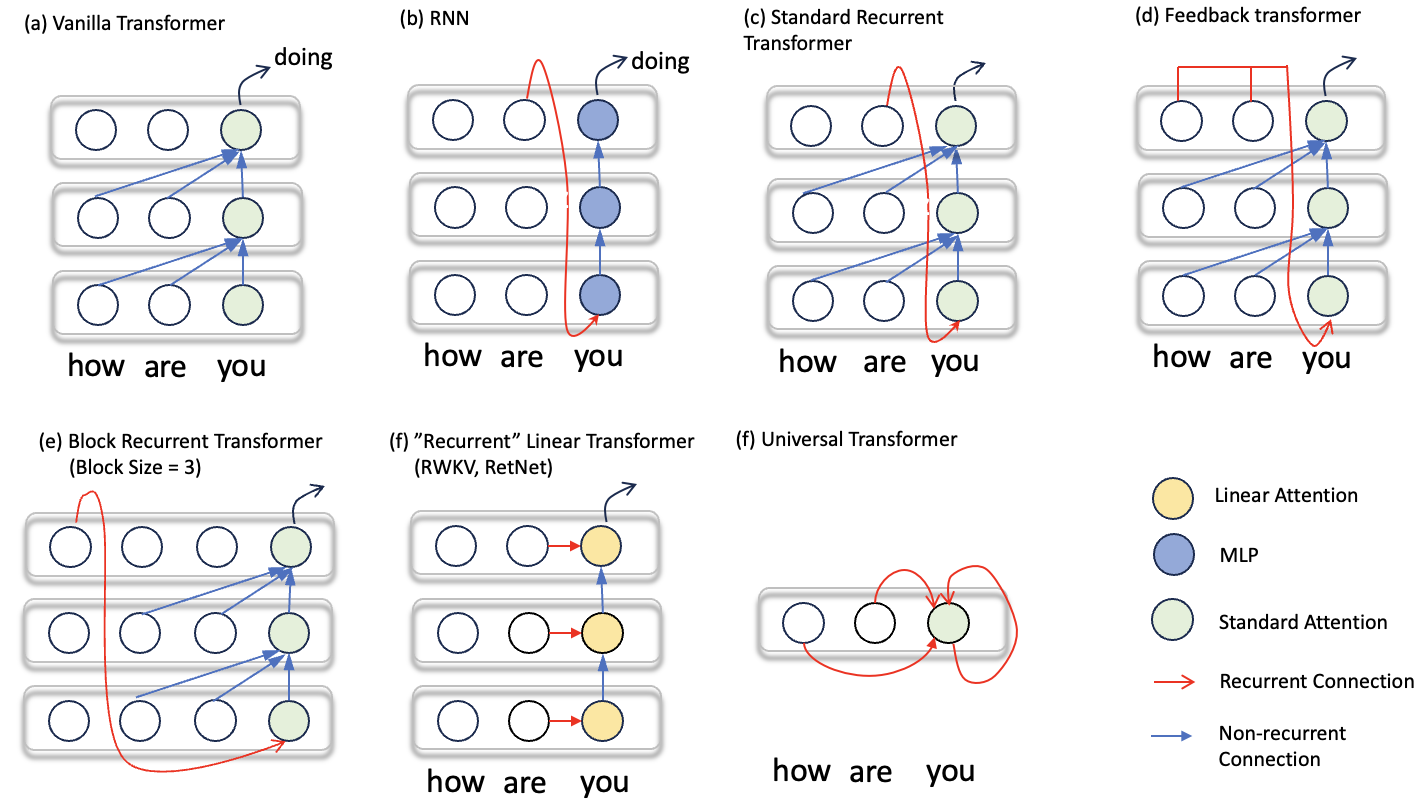} 
    \caption{Architecture diagrams of all discussed models.}
    
    \label{fig:recurTransformer}
\end{figure*}

\section{Recurrent Transformer}
Recurrence is crucial in the reasoning process, sparking extensive research into integrating recurrent features into Transformer architectures. This section explores various designs for embedding recurrence into Transformer models and proposes two categories: Recurrence-Complete (RC) and Recurrence-Incomplete (RI). Figure \ref{fig:recurTransformer} provides an overview of all discussed models. The summarized depth complexity analysis and other model properties are included in Table \ref{tab:recTsf}.

\subsection{Recurrence-Complete (RC) Models}
A model is said to be \textit{recurrence-complete} if it can represent any recurrent function as specified in Equation \ref{eq:recur}. We first illustrate how recurrence-completeness is achieved using the simplest recurrent network, RNN, and then extend this analysis to Transformer-based RC models.

\begin{figure}[htbp]
    \centering
    \includegraphics[width=0.8\linewidth]{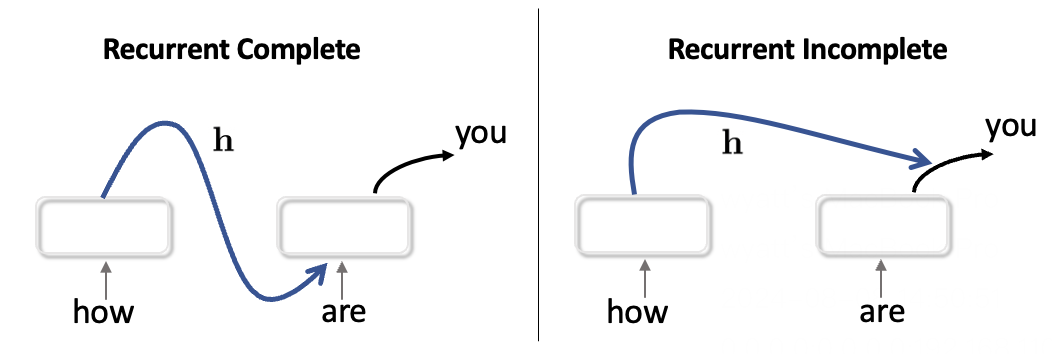} 
    \caption{A comparison between RC and RI.}
    
    \label{fig:compleness}
\end{figure}

As demonstrated in Equation \ref{eq:rnn}, RNNs model the recurrent function\footnote{One term recurrent function.} \( \mathbf{h}_t = g_\theta(\mathbf{h}_{t-1}) \) by recursively taking the previous output \( \mathbf{h} \) as the model's input (Figure \ref{fig:recurTransformer}(b) and Figure \ref{fig:compleness}, left). Given that the function \( g_\theta \), parameterized by the RNN network, incorporates both linear and nonlinear activation functions, by the Universal Approximation Theorem~\cite{cybenko1989approximation}, for any given (one term) recurrent function \( g' \), we have \( \forall \epsilon>0 \), \( \exists \theta :  |g'(\mathbf{h}) - g_\theta(\mathbf{h})| < \epsilon \). In other words, the model-encoded function \( g_\theta \) can infinitesimally  approximate or simulate any function \( g' \) such that \( \mathbf{h}_t = g'(\mathbf{h}_{t-1}) \), to an arbitrary degree of precision. Therefore, RNNs possess the capability to simulate any one-term recurrent function.

\subsection{RC Transformers}
\textbf{Standard Recurrent Transformer.} The Standard Recurrent Transformer~\cite{yang2022recurring} integrates recurrent connections of $\mathbf{h}$ with the original attention mechanism, as depicted in Figure \ref{fig:recurTransformer}c. At each time step \( t \), the computation of the first layer's key (\( \mathbf{k} \)), query (\( \mathbf{q} \)), and value (\( \mathbf{v} \)) incorporates not only the current input \( \text{x}_t \) but also the output hidden vector from the previous time step \( \mathbf{h}^{(m)}_{t-1} \):
\begin{equation}
     \mathbf{k}^{(1)}_t, \mathbf{q}^{(1)}_t, \mathbf{v}^{(1)}_t = \mathbf{W}_{k, q, v} (\text{x}_t + \mathbf{h}^{(m)}_{t-1})
\end{equation}
The subsequent layers retain the standard attention mechanism in the standard Transformer. Since each input \( \text{x}_t \) is enhanced by the previous \( \mathbf{h}^{(m)} \), the output of the final $\mathbf{h}$ at the current time step \( t \) is a function of both \( \text{x}_{1:t} \) due to the attention operations and the previous \( \mathbf{h}^{(m)}_{t-1} \) from recurrent connection, and therefore is recurrent:
\begin{equation}
    \mathbf{h}^{(m)}_{t} = g_\theta(x_{1:t}, \mathbf{h}^{(m)}_{t-1}) 
    \label{eq:stdTsf}
\end{equation}
where \( g_\theta \) represents the function embodied by the entire network.  Given the transformer's architecture consists of both linear and nonlinear layers, function \( g_\theta \) satisfies the conditions of the Universal Approximation Theorem, and therefore is Recurrence-Complete using the same argument as before.
 
\noindent\textbf{Feedback Transformer.} Instead of adding the previous output \( \mathbf{h}^{(m)}_{t-1} \) to the current input \( \text{x}_t \) as in Equation \ref{eq:stdTsf}, the Feedback Transformer~\cite{fan2020addressing} uses the attention mechanism to combine the previous $k$ terms of $\mathbf{h}^{(m)}_{t-k, t-1}$ with the current \(\text{x}_t\), as illustrated in Figure \ref{fig:recurTransformer}d. This modification improves gradient flow and optimizes the model's performance as attention allows gradient to flow through multiple optimization paths.

However, this alteration does not further improve the computational depth compared to the Standard Recurrent Transformer, as \(\mathbf{h}^{(m)}_t\) can be viewed with the same dependency as in Equation \ref{eq:stdTsf}, having a depth complexity of \(O(n)\) owing to its recurrent connection. Since \(\mathbf{h}\) is applied through an attention layer consisting of both linear and nonlinear components, following the principles of the Universal Approximation Theorem, the Feedback Transformer is Recurrence-Complete.

\noindent\textbf{Block (Recurrent) Transformer.}
The Block Transformer segments the input sequence into blocks of every \( k \) tokens, adding a standard recurrent connection only between each adjacent block. Within each block, it functions as a standard Transformer, applying attention solely among its \( k \) tokens in that segmented block. The output hidden state \( \mathbf{h}^{(m)} \) at the last token of each block is then recurrently passed to the next block along with the next \( k \) tokens as input, as illustrated in Figure \ref{fig:recurTransformer}e. Specifically, at time \( t \), \( \mathbf{h}^{(m)}_t \) is a function of both the input \( \text{x}_{t-k:t} \) within that block and the final hidden state of the previous block \( \mathbf{h}^{(m)}_{t-k-1} \), denoted as:
\begin{equation}
    \mathbf{h}^{(m)}_{t} = g_\theta(\text{x}_{t-k:t}, \mathbf{h}^{(m)}_{t-k-1})
\end{equation}
As the recurrence does not happen at each time step but every \( k \) steps, the depth complexity is only \( O(n/k) \) for a given input of length \( n \). Similar to the standard RNN and standard Recurrent Transformer, the Block Transformer is Recurrence-Complete.

\noindent\textbf{Universal Transformer.} Unlike the models discussed above, which are recurrent over time \( t \) (temporal recurrent), the Universal Transformer is recurrent over layers (depth recurrent). Specifically, unlike MLP or Transformer layers where each layer represents a different function \( g^{(i)}_\theta \) parameterized by a different weight matrix \( \mathbf{W}^{(i)} \), the Universal Transformer has a single layer. The output of this layer \( \mathbf{h} \) is recurrently recycled back as input for the same layer (Figure \ref{fig:recurTransformer}f):
\begin{equation}
    \mathbf{h}^{(i)}_{1:t} = g_\theta(\mathbf{h}^{(i-1)}_{1:t})
    \label{eq:universal}
\end{equation}
Unlike the standard Transformer, which has a fixed number of layers \( m \), the Universal Transformer can dynamically iterate the layer depth for \( T(n) \) times, with \( T \) being a function predicted by another neural network. Ideally, for tasks that are difficult and require greater depth, \( T(n) \) will be large. For tasks that are easier and can be solved with fewer layers, \( T(n) \) will be small, thus adjusting its depth complexity \( O(T(n)) \) dynamically according to need. Similarly, Equation \ref{eq:universal} conforms to the recurrent definition and is Recurrence-Complete, as function \( g_\theta \) is represented by an attention layer consisting of both linear and nonlinear components.

\subsection{Recurrent-Incomplete (RI) Models}
Some Transformer variants, though described as "recurrent," do not fully model the general recurrence function as delineated in Equation \ref{eq:recur}. These models leverage the recurrence concept to streamline complex attention computations by iterating over intermediate results. This modification avoids the need for recalculating attention from time step 1 to \( t \) at each iteration, significantly reducing the time complexity of the attention mechanism during inference and improving efficiency. However, this approach neither enhances depth complexity nor achieves genuine recurrence modeling.

Specifically, such models recurrently update previous attention aggregations and store them for the next attention computation. However, the  recurrent variable is updated through a fixed "shifting operation" rather than a learned function by the model itself (Figure \ref{fig:compleness}), thus only mimicking linear recurrent relations. 

\subsection{RI Transformer}
\textbf{RWKV.} As opposed to the standard attention mechanism, RWKV employs a modified linear attention function, defined as follows for the $i$th layer:
\begin{align}
    \mathbf{k}^{(i)}_t, \mathbf{v}^{(i)}_t &= \mathbf{W}_{k, v} \ \mathbf{h}_t^{(i-1)} \\
    \mathbf{h}_t^{(i)} &= \text{RWKVLinearAttn} ( \mathbf{k}^{(i)}_t,  \mathbf{v}^{(i)}_{1:t}) \\
     &= \frac{\sum_{j=1}^{t-1} e^{-(t-1-j)w+\mathbf{k}^{(i)}_j} \mathbf{v}^{(i)}_j + e^{u+\mathbf{k}^{(i)}_t} \mathbf{v}^{(i)}_t}{\sum_{j=1}^{t-1} e^{-(t-1-j)w+\mathbf{k}^{(i)}_j} + e^{u+\mathbf{k}^{(i)}_t}}
     \label{eq:RWKV}
\end{align}
where \( w \) and \( u \) are constant vectors.

Similar to the standard attention function, directly applying \text{RWKVLinearAttn} using vector values \( \mathbf{k} \) and \( \mathbf{v} \) is complex due to its dependence on \( \mathbf{v}, \mathbf{k} \) values from  steps \( 1 \) to \( t \). However, since RWKVLinearAttn removes the non-linear relations between pairs of \( \mathbf{k} \) and \( \mathbf{q} \) in the standard attention, \( \mathbf{h}_t^{(i)} \) can now be reformulated recursively using only the intermediate results from the \((t-1)\)-th step, significantly streamlining the function. At each time step \( t \), RWKV stores two intermediate values: \( \mathbf{a}^{(i)}_t = \sum_{j=1}^{t-1} e^{-(t-1-j)w+\mathbf{k}^{(i)}_j} \mathbf{v}^{(i)}_j \) and \( \mathbf{b}^{(i)}_t = \sum_{j=1}^{t-1} e^{-(t-1-j)w+\mathbf{k}^{(i)}_j} \), enabling the calculation of \( \mathbf{h}_t^{(i)} \) using solely \( \mathbf{a}^{(i)}_{t-1} \) and \( \mathbf{b}^{(i)}_{t-1} \) from the previous time step as follows:
\begin{align}
    \mathbf{h}_t^{(i)} = \frac{\mathbf{a}^{(i)}_{t-1} + e^{u+\mathbf{k}_t^{(i)}} \mathbf{v}_t }{\mathbf{b}^{(i)}_{t-1} + e^{u+\mathbf{k}^{(i)}_t}}
\end{align}
This way, \( \mathbf{h}_t \) is no longer dependent on values from all time steps \( 1 \) to \( t \) as in Equation \ref{eq:RWKV}, but only on values from steps \( t-1 \) and \( t \).
Values \( \mathbf{a}_t \) and \( \mathbf{b}_t \) are stored and updated at each time step for each layer \( i \) as follows:
\begin{align}
\mathbf{a}^{(i)}_t &= z \mathbf{a}^{(i)}_{t-1} + g'_\theta(\text{x}_{t}) \label{eq:a}\\
\mathbf{b}^{(i)}_t &= z \mathbf{b}^{(i)}_{t-1} + g''_\theta(\text{x}_{t}) \label{eq:b}
\end{align}
where \( z \) is a constant value \( e^{-w} \), referred to as the positional shift. The functions \( g'_\theta \) and \( g''_\theta \) are represented by the \( i \)th network layer, using the network's weights \( \mathbf{W} \) for their computations: \( g'_\theta(\text{x}_{t}) = e^{\mathbf{k}_t} \mathbf{v}_t \) and \( g''_\theta(\text{x}_{t}) = e^{\mathbf{k}_t} \). Here, the calculations for \( \mathbf{a} \) and \( \mathbf{b} \) in Equations \ref{eq:a} and \ref{eq:b} are indeed recurrent, as defined in Equation \ref{eq:recur}. Both values are recurrently derived from the previous \( \mathbf{a} \) and \( \mathbf{b} \) values output by the same layer \( i \).

The recurrent formula in Equation \ref{eq:a} for \( \mathbf{a}_t \) can be further simplified to:
\begin{align}
\mathbf{a}_t = z \mathbf{a}_{t-1} + c_{t} \label{eq:shift}
\end{align}
where \( c_t \) can be viewed as constant at each timestep since it does not depend on the recurrent variable \( \mathbf{a} \) but only on the input \( \text{x} \).

\begin{figure}[htbp]
    \centering
    \includegraphics[width=1\linewidth]{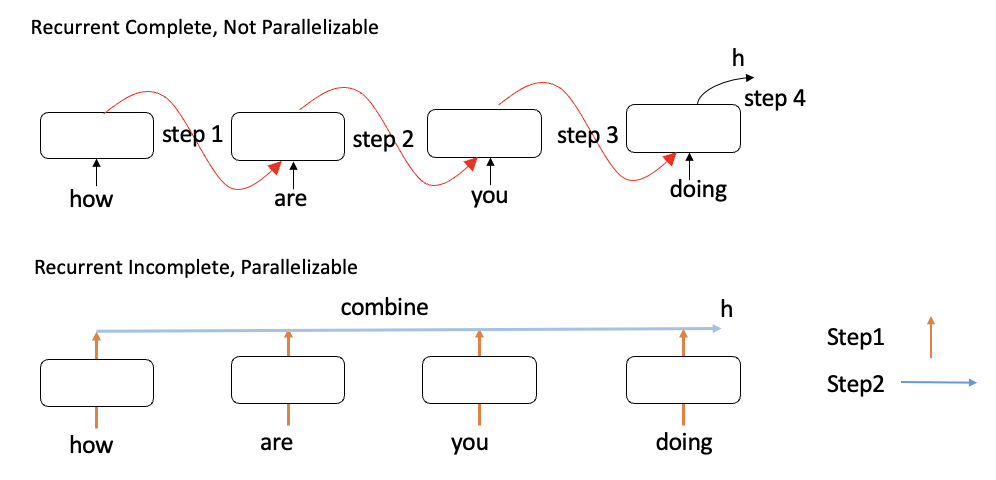} 
    \caption{How parallel training in linear-attention based Transformer achieved (bottom). In comparison, Recurrence-Complete models enforce a hard dependency between $t$ and $t-1$ steps, and sequential calculations can not be skipped. }
    
    \label{fig:parallel}
\end{figure}

Such recurrence does not represent a \emph{general} recurrent function as described in Equation 1 in Main Paper for two reasons: 
\begin{enumerate}
    \item The model-encoded function \( g'_\theta \) applies only to the input tokens \( x_t \) and not to the recurrent variable \( \mathbf{a}_{t-1} \), as shown in Figure \ref{fig:compleness} (right). Hence, the Universal Approximation Theorem does not apply to \( \mathbf{a} \) for modeling any arbitrary recurrent function. Specifically, the recurrence function \( \mathbf{a}_t \) uses a fixed shifting operation (Equation \ref{eq:shift}), with the shifted value \( c_t \) derived from \( \text{x}_t \).
    \item The model can be trained in parallel, indicating no strict dependency between values at steps \( t \) and \( t-1 \), unlike in recurrent-complete models like RNN. This parallelism is evident when expanding \( \mathbf{a}_t \) (where we fix \( z \)  to 1 for simplicity):
    \begin{align}
        \mathbf{a}_t &= \mathbf{a}_{t-1} + c_t \\
            &= (\mathbf{a}_{t-2} + c_{t-1}) + c_t \\
        &\ \ \ \ \ \cdots  \nonumber \\ 
           &= \mathbf{a}_{0} + c_{1} + \cdots + c_{t}\label{eq:shift2}
    \end{align}
    where \( \mathbf{a}_0 \) is base case setting $t$ to 0. Since each \( c_i = g'_\theta(\text{x}_i) \) depends solely on \( \text{x}_i \) and not on the previous values of  \( \mathbf{a} \), all \( \mathbf{a}_t \) values can be calculated in parallel. This parallel process is illustrated in the Figure \ref{fig:parallel}.
\end{enumerate}

Thus, while the model uses recurrent concepts to redesign the calculation of attention for enhanced inference efficiency — by avoiding recalculations from step 1 to \( t \) and by utilizing only results from step \( (t-1) \) — it does not increase depth complexity nor enable it to capture  arbitrary recurrent functions,  
rendering it an RI model.

\textbf{Linear Transformer.} Unlike RWKV, which eliminates the use of \(\mathbf{q}\) values from the standard Transformer, the Linear Transformer~\cite{katharopoulos2020transformers} preserves the usage of all \(\mathbf{k}\), \(\mathbf{q}\), and \(\mathbf{v}\) values. However, it shares the idea of using a linear function rather than the non-linear function in the standard Transformer for calculating each combination of \(\mathbf{kq}\) and \(\mathbf{v}\), as shown below:
\begin{align}
    \mathbf{k}^{(i)}_t, \mathbf{q}^{(i)}_t, \mathbf{v}^{(i)}_t &= \mathbf{W}_{k, q, v} \ \mathbf{h}_t^{(i-1)} \\
    \mathbf{h}_t^{(i)} &=  \text{LinAttn} (\mathbf{k}^{(i)}_{1:t}, \mathbf{q}^{(i)}_t,  \mathbf{v}^{(i)}_{1:t}) \\
    &= \frac{\sum_{i=1}^{t} {\phi(\mathbf{q}_t^{(i)}) \phi(\mathbf{k}_i^{(i)}) \mathbf{v}_i^{(i)}}}{\sum_{i=1}^{t} {\phi(\mathbf{q}_t^{(i)}) \phi(\mathbf{k}_i^{(i)})}}
    \label{eq:linearTsf}
\end{align}
where \(\phi(\text{x})\) is independently applied to each value in vectors \(\mathbf{q}\) and \(\mathbf{k}\) before linearly multiplying them together. Similar to RWKV, Equation \ref{eq:linearTsf} can now be computed using solely the intermediate values \(\mathbf{a}_{t-1}\) and \(\mathbf{b}_{t-1}\) from time step \( t-1 \), rather than using all \(\mathbf{q}\), \(\mathbf{k}\), and \(\mathbf{v}\) values from step 1 to \( t \):
\begin{equation}
     \mathbf{h}_t^{(i)} = \frac{\phi(\mathbf{q}_t^{(i)}) \mathbf{a}^{(i)}_{t-1}}{\phi(\mathbf{q}_t^{(i)}) \mathbf{b}^{(i)}_{t-1}}
\end{equation}
with \(\mathbf{a}\) and \(\mathbf{b}\) recurrently computed as follows:
\begin{align}
\mathbf{a}^{(i)}_t &= \mathbf{a}^{(i)}_{t-1} + g'_\theta(\text{x}_{t})  \\
\mathbf{b}^{(i)}_t &= \mathbf{b}^{(i)}_{t-1} + g''_\theta(\text{x}_{t})
\end{align}
where \( g'_\theta(\text{x}_{t}) = \phi(\mathbf{k}_t^{(i)}) \mathbf{v}_t^{(i)} \) and \( g''_\theta(\text{x}_{t}) = \phi(\mathbf{k}_t^{(i)}) \), computed using the model weight \(\mathbf{W}\). Similar to RWKV, \(\mathbf{a}\) and \(\mathbf{b}\) are recurrently computed with a shifted value rather than computed using model weights, so the Universal Approximation Theorem does not apply to the recurrent variable of \(\mathbf{a}\) and \(\mathbf{b}\) (Figure \ref{fig:compleness} right). Therefore, the Linear Transformer represents another instance in the RI class.

\subsection{Parallel Training of RC and RI Models}
As we can see, Recurrence-Complete models do not support either parallel training nor inference due to the recurrent formula \(\mathbf{h}_t = g(\mathbf{h}_{t-1})\), which enforces a hard dependency; the result at time \( t \) cannot be computed until \(\mathbf{h}_{t-1}\) is obtained, as shown in the top part of Figure \ref{fig:parallel}. However, Linear-Attention based Recurrence-Incomplete models allow for parallel training because the recurrence in their design \(\mathbf{a}_t = g(\mathbf{a}_{t-1}) = \mathbf{a}_{t-1} + c\) is only a shifting operation (Equation \ref{eq:shift}), and such an operation is associative and commutative. Specifically, to obtain \(\mathbf{a}_t\), we shift \(\mathbf{a}_0\) with values \(c_1, c_2, \cdots, c_t\) (Equation \ref{eq:shift2}). The associative and commutative properties allow us to shift the value of \( c \) in any order without strict dependency. During training, all the shifted values \(c_1, c_2, \cdots, c_t\) can be calculated in parallel since each \(c_i\) is computed independently by applying the model to the corresponding input: \(c_i = \mathbf{W}\text{x}_i\). Therefore, shifting can be done in parallel with all values of $c$ obtained, as demonstrated in the bottom part of Figure \ref{fig:parallel}.

\subsection{No Free Lunch for Parallelism}

We propose a "No Free Lunch" rule for parallel computing in neural models: parallel training is a must trade-off for Recurrent-Completeness, and both cannot be achieved simultaneously. Specifically, a true recurrent (RC) model cannot be parallelized during either inference or training, as the computation of $\mathbf{h}_{t+1}$ strictly depends on $\mathbf{h}_{t}$ in a sequential manner. 

This can be proven by contradiction. Assume a true \textbf{recurrent} model can be trained or inferred in parallel. Then the acquisition of $\mathbf{h}_{t+1}$ can occur at the same time as $\mathbf{h}_{t}$, meaning that $\mathbf{h}_{t}$ is not a necessary dependency for $\mathbf{h}_{t+1}$. This implies that $\mathbf{h}_{t+1}$ could be computed using some other variable, say $\mathbf{v}$, which is independent of $\mathbf{h}_{t}$. Consequently, this model would not be \textbf{recurrent}, as $\mathbf{h}_{t+1}$ can be expressed as a function of solely $\mathbf{v}$, $g(\mathbf{v})$, contradicting our initial assumption of the model being recurrent. 

Linear Transformers' $\mathbf{h}$ can be understood in this way, since $\mathbf{h}$ can be fully expressed using $\mathbf{x}$ rather than the previous $\mathbf{h}$, as in RC models. Therefore, both recurrence and parallel training cannot be attained simultaneously. Recurrent Neural Networks (RNNs) sacrifice parallel training for recurrent connections, while Transformers trade recurrence for parallelism.

\section{Conclusion}
In this work, we analyzed the distinct roles of autoregression and recurrence in a model's reasoning process, demonstrating that recurrence is crucial for boosting computational depth. We explained that CoT approximates recurrence in Transformer-based autoregressive LLMs from a computational standpoint. Lastly, our analysis of recurrence completeness highlights the importance of choosing the right structure for different tasks, as some "recurrent" structures aim to increase inference speed rather than depth complexity. Our findings offer insights for designing new Transformer-based models with enhanced computational and reasoning capabilities.

\bibliography{aaai25}
u re\newpage
\newpage
\section*{Appendix}

\section*{Experiment}
\subsection*{Controlled Experiment Designs}
Our goal is to demonstrate that recurrence can enhance the depth of reasoning in neural models. To achieve this, we carefully control our experiments to minimize the influence of other factors that could affect the model's performance. 
Please note that the version of GPT-4 used in this experiment was released before November 2023. Results may vary slightly due to potential changes in the version and updates to the model.

In previous discussions, we've highlighted how tokenization can significantly impact a model's failure on certain tasks. To mitigate this, we’ve designed alternative task formats that rule out the effects of tokenization. Additionally, since optimization is often imperfect, large language models (LLMs) can struggle with long-context information retrieval and may produce hallucinations as context length increases. This, in turn, can negatively affect testing accuracy, as models often fail to reference the original task instances and values during extended reasoning steps. While these factors are critical in real-world LLM applications, they are distracting for our experimental purposes, which focus on the model's architecture and computational ability rather than optimization effects.

Therefore, we limit our experimental task lengths to under 20 elements and we sample lengths when generating task instances (except for list reversing where we sample length from 30 to 40). This threshold was determined through preliminary analysis, which showed that CoT processes become unmanageably long and prone to non-computability-related errors when task instances exceed 20 steps. When conversation length increases significantly, models tend to split outputs across multiple sessions,  complicating accurate information retrieval due to imperfect optimization. By limiting length, we stay within a manageable context length, minimizing the aforementioned issues while still being able to demonstrate the difference between CoT and non-CoT in reasoning process.

Modern LLMs are fine-tuned to perform Chain of Thought reasoning by default. When prompted, they typically engage in step-by-step intermediate reasoning before providing an answer. For LLMs without CoT, we therefore explicitly forbid the use of CoT in our prompts, instructing the model to "Give a direct answer without steps." This ensures that reasoning occurs solely within the hidden representations of the Transformer network, avoiding the vector-to-string conversion discussed in the CoT process.

Finally, to address the inherent variability in LLM generation process, which involves statistical sampling, we conduct multiple trials for each task instance. We generate 50 task instances per task and perform reasoning three times independently for each instance. An answer is considered correct if at least one of the three prompts yields the correct result. This approach aligns with the experimental settings used in baseline expert models~\cite{deletang2022neural}, where models are trained 10 times for each task, and the best-performing model is used for testing. This allow us to focus on the upper bound of performance rather than average performance, ensuring that errors due to randomness are minimized.

\subsection*{Tasks}
The tasks are designed to assess the model's computability rather than its "intelligence", following the previous work's~\cite{deletang2022neural} task design with modifications for LLMs. This means that all tasks involve simple rule iterations and memory access rather than complex algorithm design. However, successfully solving these tasks requires the model's architecture and memory system to meet or exceed the complexity level needed for each task. Below, we provide a detailed description of each task design, along with sample inputs and outputs. Lengths of all instances $n$ are sampled from 10 to 20. 

We use three tasks in the Regular (R) class:

\begin{enumerate}
    \item \textbf{Modular Arithmetic:} Given a sequence of \( n \) numbers and operations (+, -), compute the result modulo 5. For example, the input \( 1 + 3 - 2 \) should yield 2.

    \item \textbf{Parity Check:} Given a list containing the words "apple" and "banana," determine if the word "apple" appears an even number of times. For example, the input \texttt{("apple", "apple", "banana")} yields \texttt{True}.

    \item \textbf{Cycle Navigation:} Given a list of actions ("forward," "backward," "stay"), determine the final position in a 5-state cycle, starting from state 1. For example, the input \texttt{("forward", "forward", "backward")} will result in state 1. This task is equivalent to Modular Arithmetic.
\end{enumerate}

We use three tasks in the Context-Free (CF) class:

\begin{enumerate}
    \item \textbf{Stack Manipulation:} Given a list of values (fruit names) representing a stack, and a sequence of \( n \) actions, compute the resulting stack. For example, applying the actions \texttt{(pop "apple", push "peach")} to the stack \texttt{("grape", "banana", "apple")} results in \texttt{("grape", "banana", "peach")}.
    
    \item \textbf{Reverse List:} Given a list of fruit names, reverse the list.
    
    \item \textbf{Modular Arithmetic (Complex):} Given an arithmetic expression with \( n \) operations, calculate the result modulo 5. For example, \(((3 + 4) - 1) \times (2 + (1 - 2))\) yields 1.
\end{enumerate}

We use four tasks in the Context-Sensitive (CS) class:

\begin{enumerate}
    \item \textbf{Odd First:} Given a list of fruits, extract all fruits at odd positions, followed by fruits at even positions. For example, \texttt{("apple", "grape", "banana", "peach")} yields \texttt{("apple", "banana", "grape", "peach")}.
    
    \item \textbf{Addition:} Given two large numbers with \( n \) digits, calculate the sum.
    
    \item \textbf{Multiplication:} Given two large numbers with \( n \) digits, calculate the product.
    
    \item \textbf{Sorting:} Given a list of numbers, sort them using the insertion sort algorithm.
\end{enumerate}

\subsection*{Case Studies}

In this section, we provide a detailed analysis of how Chain of Thought (CoT) models recurrence by converting $\mathbf{h}$ into text and then back into $\mathbf{h}$ for recurrent reasoning, using the model's output on various task instances. We also demonstrate through case studies how non-CoT-based large language models (LLMs) fail to achieve the same.

We present the output when GPT-4 is used to solve tasks such as counting, sorting, list reversing, and number addition. As shown in Figure \ref{fig:case_sort} at the end of the document, CoT sorts a list by repeatedly converting the partially sorted list encoded in $\mathbf{h}$ into natural language, then converting it back to $\mathbf{h}$ for further computation. This approach allows sorting to be performed without relying solely on the model's layer depth, which is fixed at $O(1)$. Instead, the computation can be extended through vector-string conversions, enabling the reasoning process to continue arbitrarily long using autoregressive generation, thus enhancing the depth of reasoning.

\begin{figure*}[htbp]
    \centering
    \includegraphics[width=0.7\linewidth]{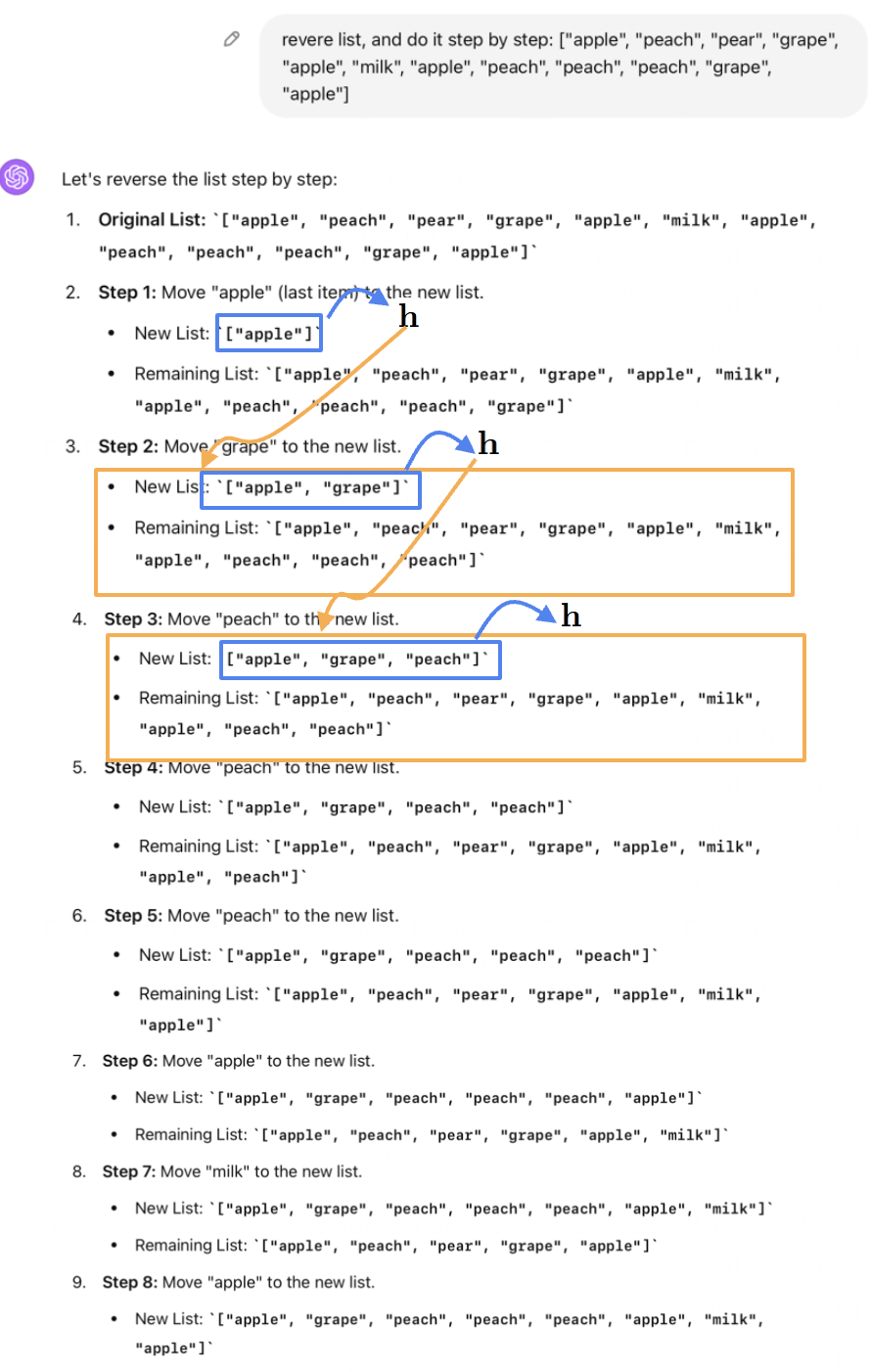} 
    \caption{Case study of list reversing. CoT effectively models recurrent computation by iteratively converting the computed result from $\mathbf{h}$ to text $\mathbf{o}$. This text $\mathbf{o}$ is then read back into a vector form $\mathbf{h}$ for the next computation. The process of $\mathbf{h} \rightarrow \mathbf{o}$ is represented in yellow, where the newly calculated list in $\mathbf{h}$ is converted into natural language list tokens. The process of $\mathbf{o} \rightarrow \mathbf{h}$ is represented in blue, where the description of the current computation, or partial list, is converted back into the vector $\mathbf{h}$ for subsequent computation.}

    \label{fig:case_reverse}
\end{figure*}

\begin{figure*}[htbp]
    \centering
    \includegraphics[width=0.8\linewidth]{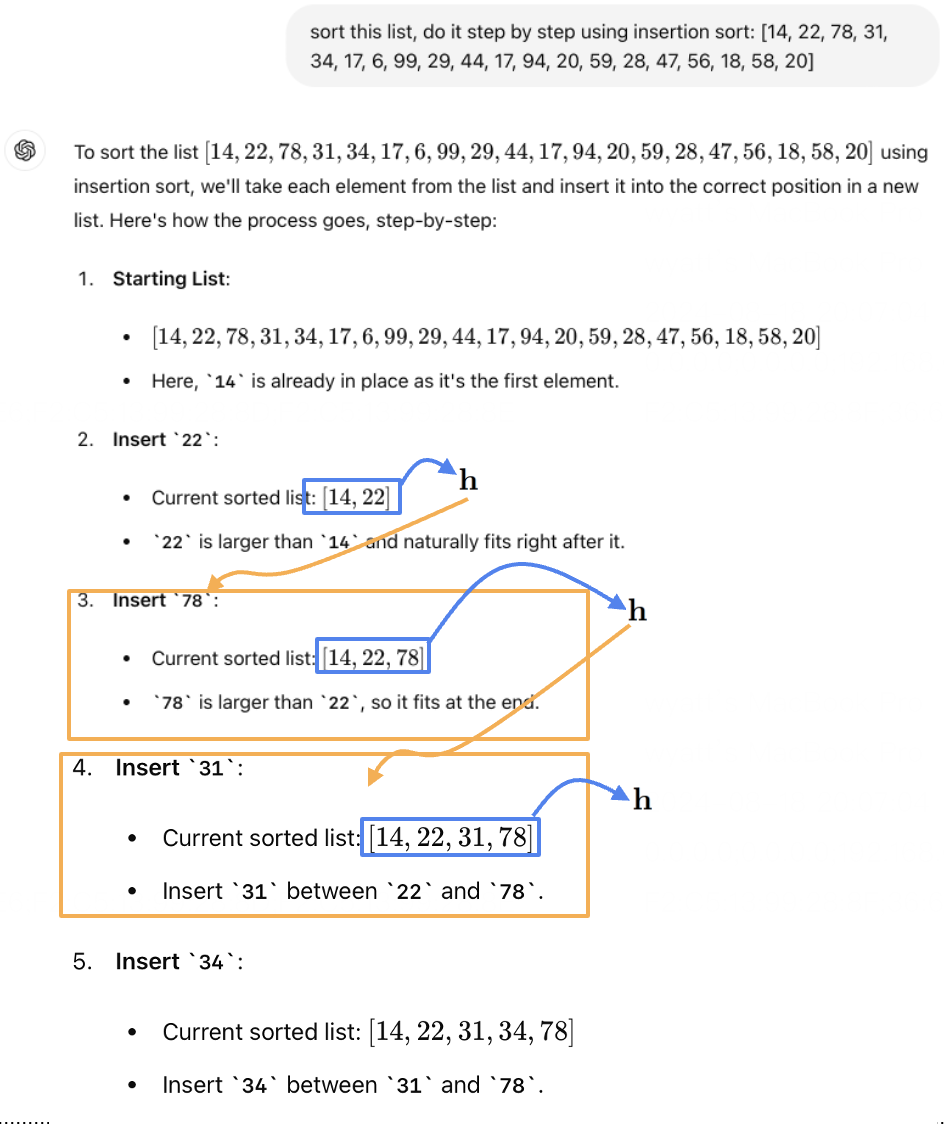} 
    \caption{Case study of sorting. The process $\mathbf{h} \rightarrow \mathbf{o}$ is represented in yellow, where the sorted list in $\mathbf{h}$ is converted into natural language list tokens. The process $\mathbf{o} \rightarrow \mathbf{h}$ is represented in blue, where the description of the current computation, or partially sorted list, is converted back into the vector $\mathbf{h}$ for the next computation.}

    \label{fig:case_sort}
\end{figure*}

\begin{figure*}[htbp]
    \centering
    \includegraphics[width=\linewidth]{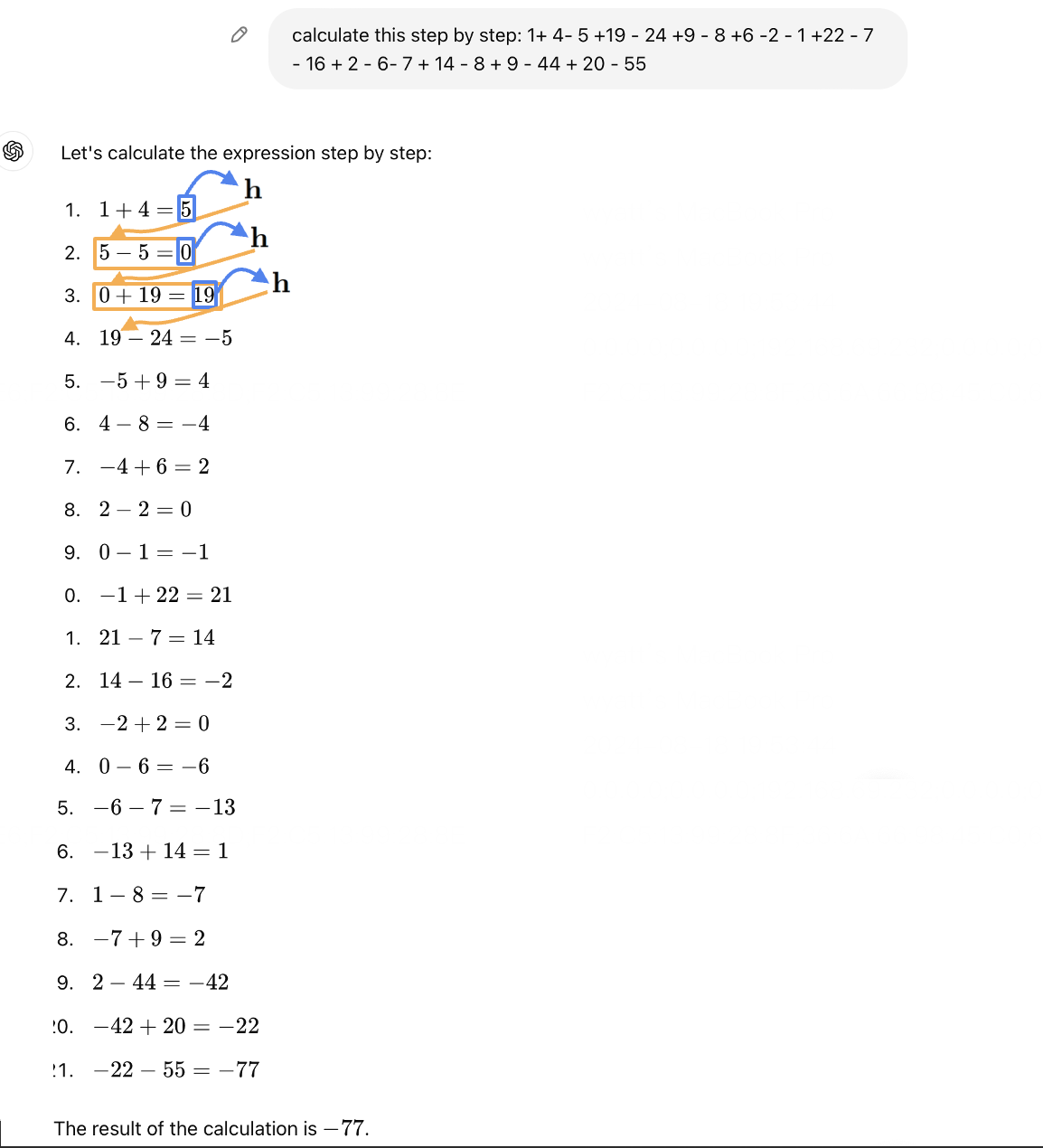} 
    \caption{Case study of addition. The process $\mathbf{h} \rightarrow \mathbf{o}$ is represented in yellow, where the current addition result in $\mathbf{h}$ is converted into natural language. The process $\mathbf{o} \rightarrow \mathbf{h}$ is represented in blue, where the current added value is converted back into the vector $\mathbf{h}$ for the next computation.}

    \label{fig:case_add}
\end{figure*}

\begin{figure*}[htbp]
    \centering
    \includegraphics[width=0.9\linewidth]{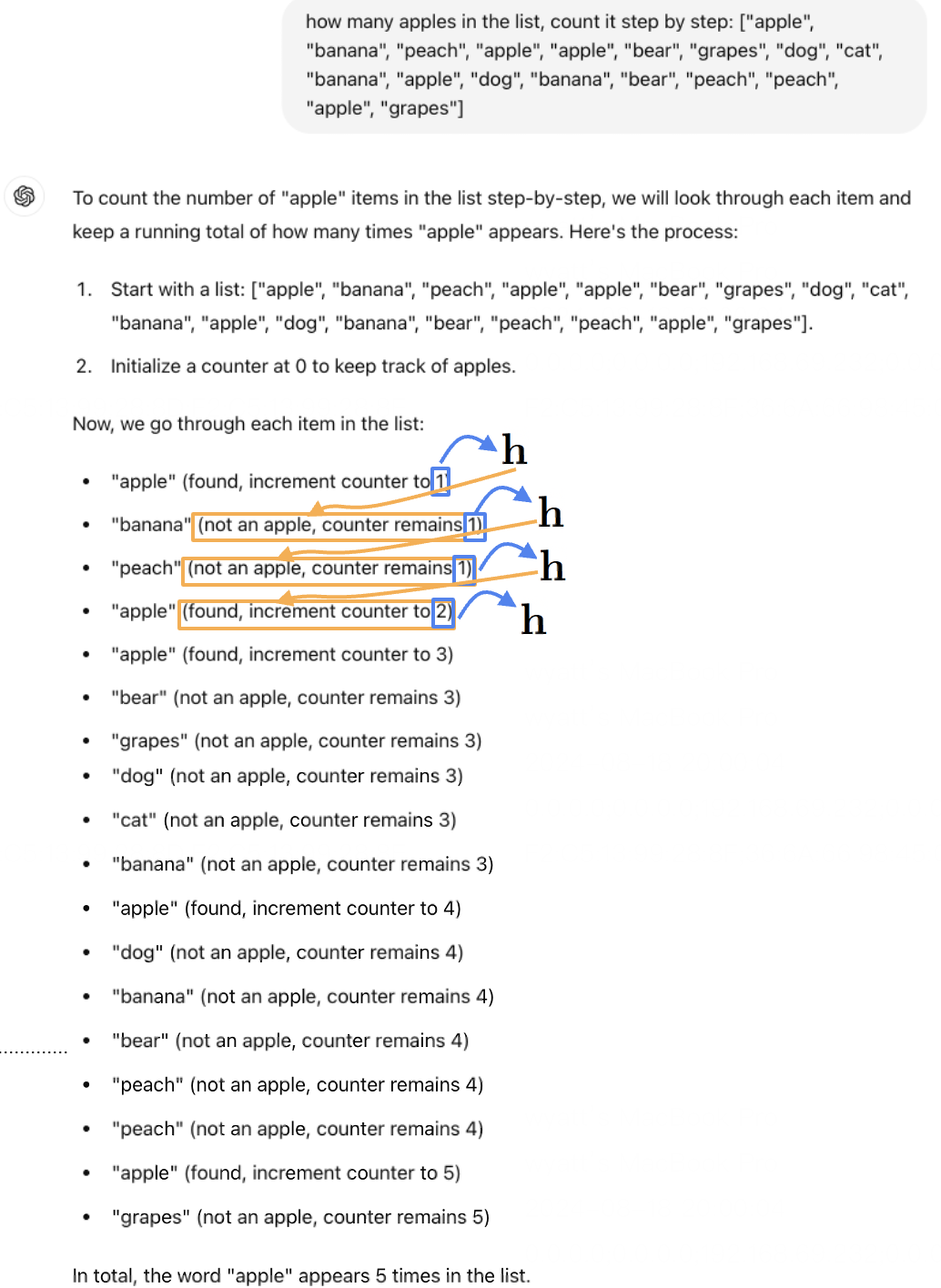} 
    \caption{Case study of counting using CoT. }
    \label{fig:case_count}
\end{figure*}

\begin{figure*}[htbp]
    \centering
    \includegraphics[width=0.9\linewidth]{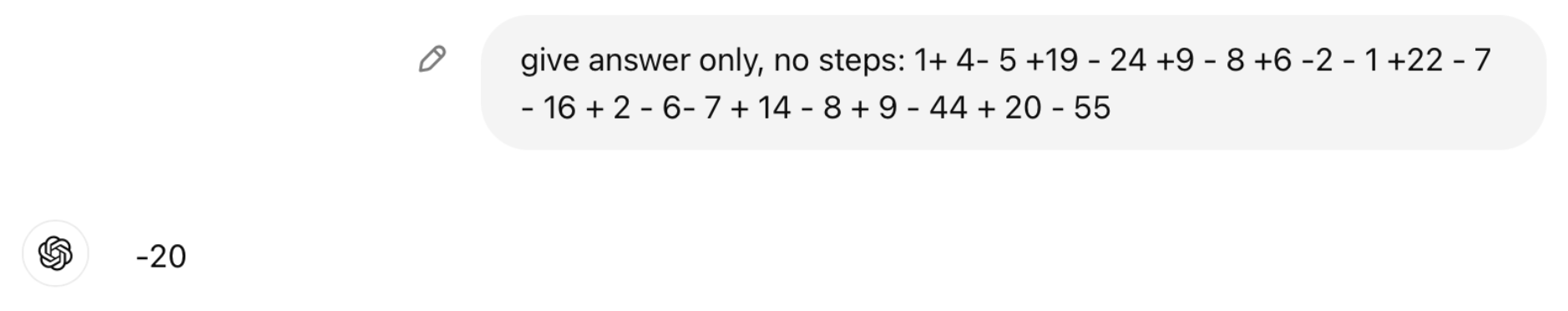} 
    \caption{Case study of addition without using CoT. Answer is incorrect.}
    \label{fig:case_add_non}
\end{figure*}

\begin{figure*}[htbp]
    \centering
    \includegraphics[width=0.9\linewidth]{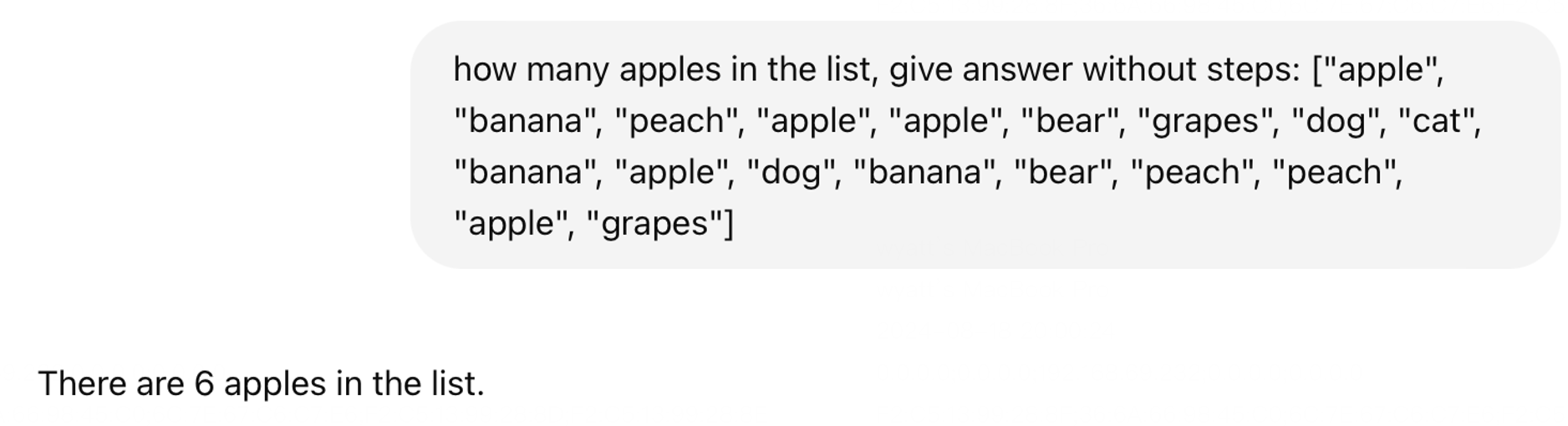} 
    \caption{Case study of couting without using CoT. Answer is incorrect.}
    \label{fig:case_count_non}
\end{figure*}

Similarly, case studies on counting (Figure \ref{fig:case_count}), list reversing (Figure \ref{fig:case_reverse}), and addition (Figure \ref{fig:case_add}) demonstrate the same process of resuming reasoning and boosting computational depth. In contrast, without intermediate results and relying only on the Transformer's inherent layer-wise depth, tasks like counting and addition cannot be completed (as shown in Figures \ref{fig:case_add_non} and \ref{fig:case_count_non}).

\newpage

\newpage

\newpage

\newpage
\newpage

\end{document}